\documentclass[journa,10ptl]{IEEEtran}
\usepackage[nocompress]{cite}
\usepackage{amsmath,amssymb,amsfonts}
\usepackage{algorithmic}
\usepackage{algorithm}
\usepackage{xcolor}
\usepackage{textcomp}
\usepackage{graphicx}
\usepackage{epsfig}
\usepackage{subfig}
\usepackage{hyperref}
\usepackage[compatibility=false]{caption}

\def\BibTeX{{\rm B\kern-.05em{\sc i\kern-.025em b}\kern-.08em
T\kern-.1667em\lower.7ex\hbox{E}\kern-.125emX}} 

\begin{document}

\title{\huge Design and Analysis of Robust Adaptive Filtering with the Hyperbolic Tangent Exponential Kernel M-Estimator Function for Active Noise Control}
\author{Iam Kim de S. Hermont, Andre R. Flores and Rodrigo C. de Lamare
\thanks{Department of Electrical Engineering, PUC-Rio, Rio de Janeiro, Brazil. Emails:  
iamkim@aluno.puc-rio.br, andre\_flores@esp.puc-rio.br, delamare@puc-rio.br}} 

\maketitle

\begin{abstract}
 In this work, we propose a robust adaptive filtering approach for active noise control applications in the presence of impulsive noise. In particular, we develop the filtered-x hyperbolic tangent exponential generalized Kernel M-estimate function (FXHEKM) robust adaptive algorithm. A statistical analysis of the proposed FXHEKM algorithm is carried out along with a study of its computational cost. {In order to evaluate the proposed FXHEKM algorithm, the mean-square error (MSE) and the average noise reduction (ANR) performance metrics have been adopted.} Numerical results show the efficiency of the proposed FXHEKM algorithm to cancel the presence of the additive spurious signals, such as \textbf{$\alpha$}-stable noises against competing algorithms.
\end{abstract}
\begin{IEEEkeywords}
Adaptive filtering, robust techniques, active noise cancellation, nonlinear systems, impulsive noise.
\end{IEEEkeywords}


\section{Introduction}

Signal processing applications suffer from the effects of undesired acoustic signals, known as \textit{noise}, which come from different sources and heavily degrade the general operation of digital signal processing systems. To ensure an appropriate performance, noise reduction (NR) approaches are fundamental. In order to cancel the noise, diverse electronic systems capable of handling different sources of interference, vibration, and reverberation have been developed over the last decades \cite{ref1}.  In particular, the research field known as active noise control (ANC) has emerged as a powerful set of tools that deal with the problem of canceling the effects of noise and has become one of the most important research fields in signal processing. 

In ANC systems, the suppression of the noise is achieved by introducing an “antinoise” wave through an appropriate segment of secondary sources \cite{ref25, ref4}. Specifically, an adaptive algorithm computes and models the ``antinoise''. For example, with the standard least-mean square (LMS) adaptive algorithm, it is possible to learn about the signal patterns through adaptive processing \cite{ref2,ref25,aifir,jio,jiodoa,jiomimo,jiols,saalt,dce,jidf,rrmser,rrlrd,dynovs}.  Then, an electro-mechanical-acoustic system propagates the “antinoise” wave, which has the same amplitude as the undesired noise but with inverse phase, thus achieving noise cancellation in the environment of interest \cite{ref25,ref5}. 

By combining the classic LMS  adaptive algorithm with another renowned system approach that uses an auxiliary path that is called filtered-X algorithm  \cite{ref8,ref2}, a powerful combination known as FX-LMS algorithms has been devised. The use of the FX-LMS algorithm for ANC applications is rather simple and stable for linear systems and Gaussian noise \cite{ref17} and, for this reason, it has been used for decades for many ANC applications and developments.  However, this approach does not work well in the presence of non-Gaussian noise, and the accuracy of the LMS usually degrades. Thus, a series of adaptive algorithms have been developed to solve the ANC problem treating a band of complex issues involving linear and nonlinear systems, different types of noise, and special conditions.

Specific cases where the performance of the FX-LMS  algorithm is degraded include non-linear structures (for instance chaotic or unpredictable) and non-Gaussian noises (such as impulsive noise). Such scenarios require one to resort to approaches that yield feasible treatment of the physical features of the signal(s) and system(s) involved \cite{ref6}. In general, to deal with this type of problem many different concepts were devised \cite{ref7}. For instance, the well-known variable step-size technique \cite{ref21} has been employed in ANC systems. Other algorithms include the recursive least-squares (RLS) \cite{ref22}, stochastic gradient descent, Newton-type and \textit{q}-gradient \cite{ref18}, the least mean $p$-power \cite{ref23} and heuristic algorithms such as Particle swarm optimization \cite{ref24}. Another cost-effective strategy is to use modified objective functions to develop robust algorithms \cite{l1stap,locsme,okspme,lrcc,rapa}. {To this end, different methods like the correntropy criterion \cite{ref8,WANG2025}, hyperbolic trigonometric functions \cite{ref9}, functional link neural networks \cite{Ye2023,Ye2024} and an exponential conjugated version \cite{ref10}, deep learning-based algorithms \cite{deepanc1,deepanc2,deepanc3} that are more costly than standard adaptive techniques, the M-estimator method \cite{ref11,dme} have been investigated.} The reader is referred to the surveys in \textit{et al} \cite{ref6,ref7} for a comprehensive account of the contributions in the last decade or so.


Preliminary results of this work appeared in our conference publication \cite{Kim2024}. This work presents the development and a statistical analysis of the filtered-x hyperbolic tangent exponential generalized Kernel M-estimate function (FXHEKM) robust adaptive algorithm for active noise control applications in the presence of impulsive noise. The proposed FXHEKM robust adaptive algorithm relies on the hyperbolic tangent exponential and the M-estimate in the objective function and employs a stochastic gradient recursion equipped with a fixed step size to adjust its parameters. A statistical analysis of the proposed FXHEKM algorithm is carried out along with an evaluation of its computational complexity. The statistical analysis provides conditions for the stability of the FXHEKM algorithm and derives a theoretical formula to predict its MSE at steady state. {Numerical results evaluate the performance of FXHEKM against competing robust adaptive algorithms for Gaussian and non-Gaussian noises. }

The manuscript is organized as follows: The system model and the problem formulation are presented and described in detail in Section \ref{sys_prob}. The mathematical derivation of the proposed FXHEKM algorithm is described in Section \ref{hekm} along with its complexity analysis. A statistical analysis of the FXHEKM algorithm is carried out in Section IV, which provides stability conditions and a formula to predict the MSE error of the FXHEKM at steady state. In Section \ref{sim_results}, the proposed method is compared with other classical and robust approaches for performance evaluation and validation. Lastly, the conclusions of this work are brought in Section V.

\section{System model and problem formulation}
\label{sys_prob}
In this section, we present the system model of an ANC system and formulate the problem that we are interested in solving.

\subsection{System model}

\begin{figure}[ht]
\centerline{\includegraphics[scale=.2]{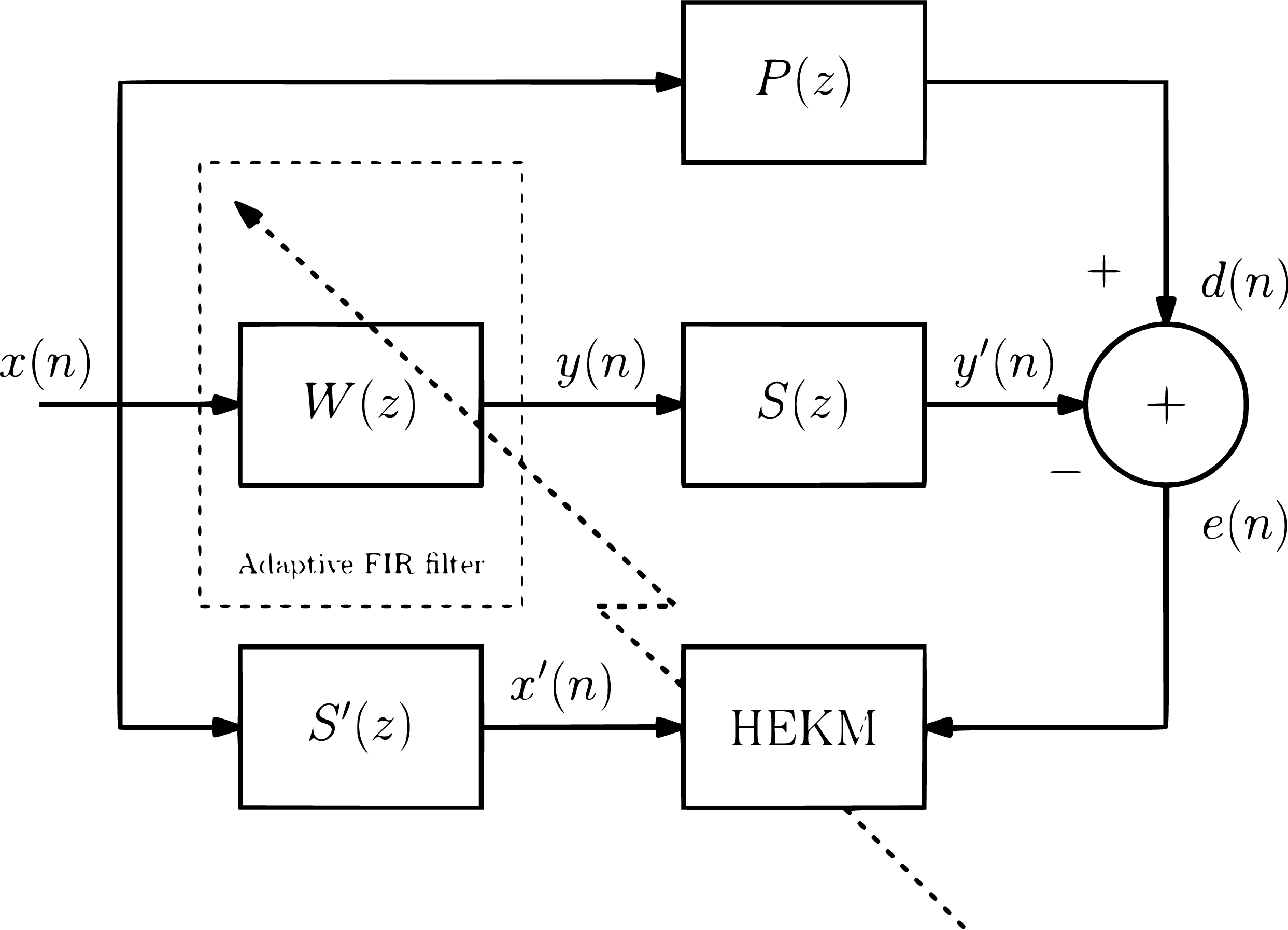}}
\caption{Block diagram of an ANC system.}
\label{fig1}
\end{figure}

Consider an Active Noise Cancellation problem, where we try to suppress the noise contaminating the environment \cite{ref4}. This task is performed by an adaptive ANC system, as shown in Fig. \ref{fig1}. The variable $x(n)$ describes the input signal; $y(n)$ represents the output signal vector; $e(n)$ is the residual error estimation of the system in the procedure of computing the value of the error to update the adaptive algorithm. 
The calculation of these variables is described in detail in the following section. Note that in this ANC system, the input signal is used directly in the error estimation and the calculation of the filter response. 

\subsection{ANC processing and problem formulation}

The components of the model crossed by a dashed line can be defined as an adaptive filter, consisting of: \textit{i}) a digital filter (here we consider a finite impulse response (FIR) structure) to develop the signal processing and \textit{ii}) the adaptive algorithm that performs the weight update of this filter. We can define the input vector at discrete time \textit{n} as
\begin{equation}
\begin{aligned}
    \mathbf{x}(n) \equiv \left[ x(n) \ \ x(n-1) \ \ ... \ \ x(n - L + 1) \right]^{T},
    \label{x}  
\end{aligned}
\end{equation}
and similarly, the weight vector of the adaptive filter as
\begin{equation}
    \mathbf{w}(n) \equiv \left[ w_0(n) \ \ w_1(n) \ \ ... \ \ w_{L - 1}(n) \right]^{T},
    \label{w}
\end{equation}
where \textit{T} denotes the transpose operator. The signal related to the estimation of input noise $x(n)$ of the filtered output can be expressed by the vector operation given by 
\begin{equation}
\begin{aligned}
    y(n) &= \mathbf{w}(n)^{T}\mathbf{x}(n). 
    \label{y}
\end{aligned}
\end{equation}
Then, the value of the residual error in the system to be computed is defined as the difference between the desired signal and the filtered output, expressed by   
\begin{equation}
\begin{aligned}
    e(n) &= d(n) - y'(n) \\ 
    &= p(n) * x(n) + v(n) - s(n) * (\mathbf{w} (n)^{T}\mathbf{x}(n)) \\
    &= d(n) - \mathbf{w} (n)^{T}\mathbf{x'}(n) ,
    \label{e}
\end{aligned}
\end{equation}
where $d(n)$ is the desired signal and $y'(n)$ is the noise input estimate computed by the adaptive filter. The parameter $v(n)$ models the measurement error, and $p(n)$ and $s(n)$ are the impulse responses at time $n$ of the primary $P(z)$ and the secondary paths $S(z)$, respectively, and * denotes the operation of linear convolution.
The signal vector $\mathbf{x}'(n)$ is the well-known \textit{filtered}-X response from the input of the ANC system. The problem that we are interested in solving is how to design a robust adaptive algorithm for ANC systems that can mitigate impulsive noise effects.

\section{Proposed FXHEKM Algorithm}
\label{hekm}


In this section, we detail the derivation of the proposed FXHEKM algorithm and carry out an evaluation of its computational complexity analysis and competing algorithms. 

\subsection{Derivation of FXHEKM}
\label{AA}

The research that led to this work aimed to find a robust function that can obtain an attractive trade-off between fast convergence with the lowest MMSE level, considering the operation under both Gaussian and impulsive signals. In what follows, we initially define the objective function of the FXHEKM algorithm. 

\begin{figure*}
\centering
\subfloat[]
{\includegraphics[width=0.475\linewidth]{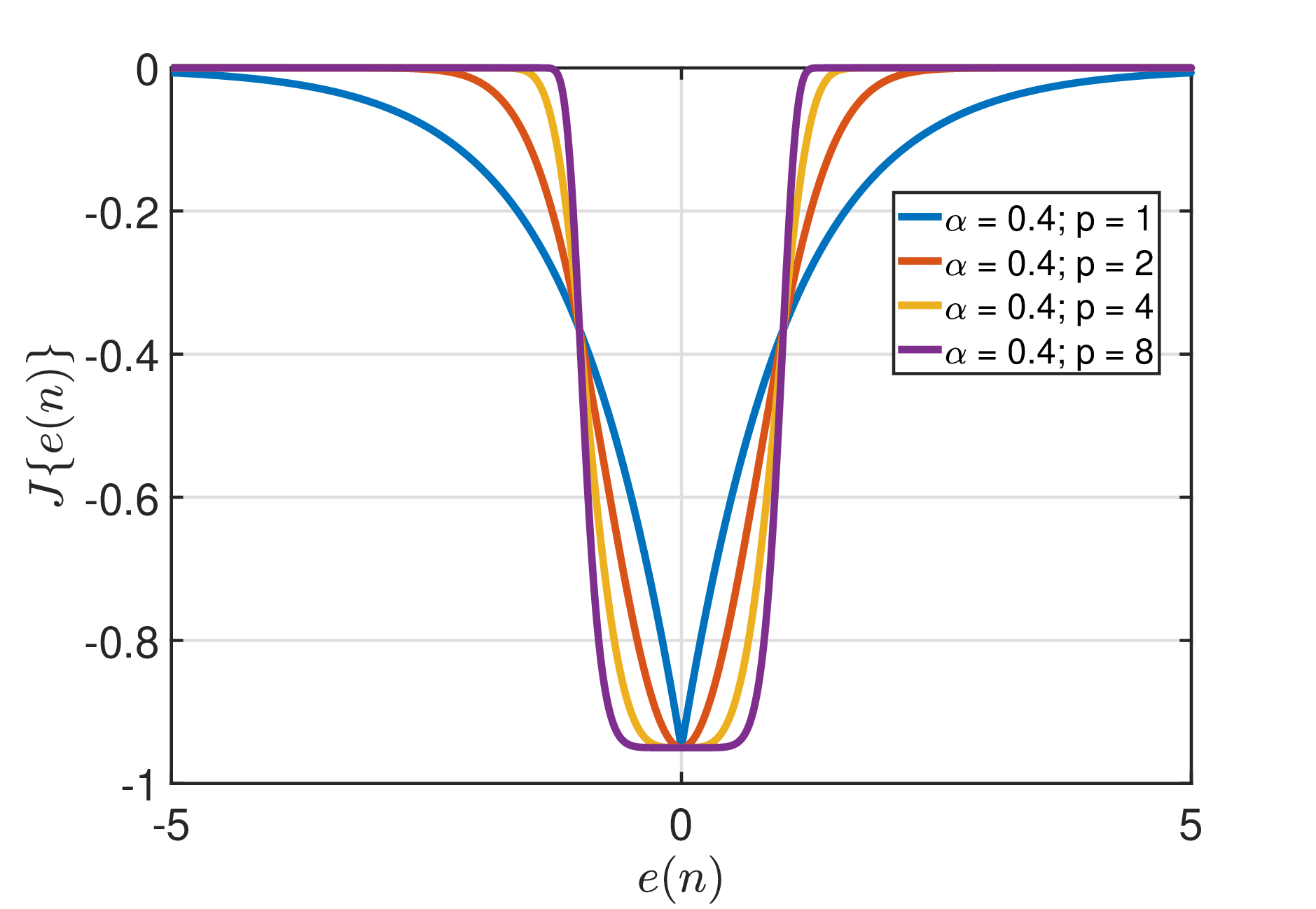}
\label{objective function p}}
\hfill
\subfloat[]
{\includegraphics[width=0.475\linewidth]{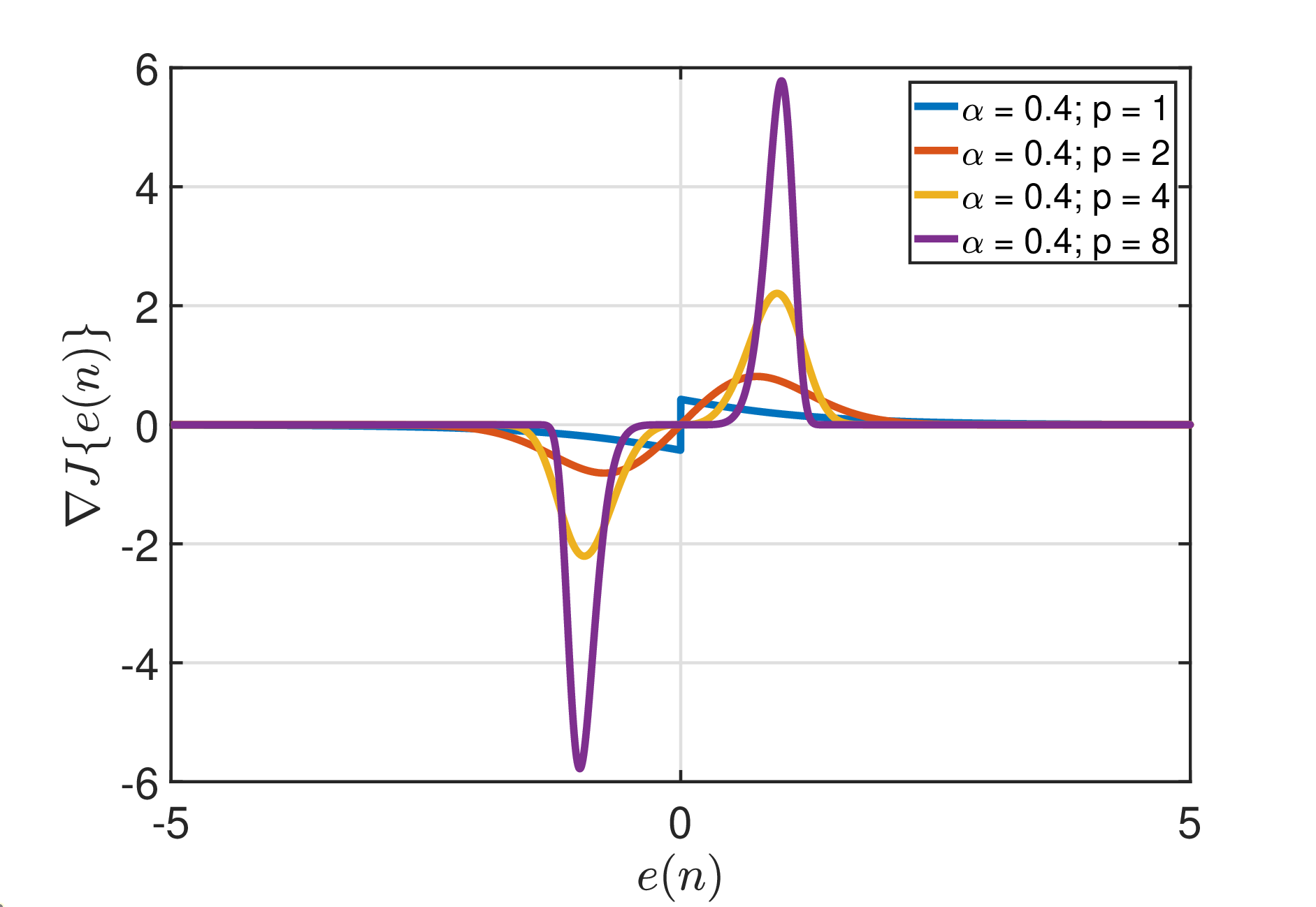}}
\label{objective function p derivative}
\vspace{-0.35em} 
\caption{\small Effect of $p$ over {(a)} the objective function and {(b)} its derivative.}
\label{objt fun p}
\end{figure*}

\begin{figure*}
\centering
\subfloat[]
{\includegraphics[width=0.475\linewidth]{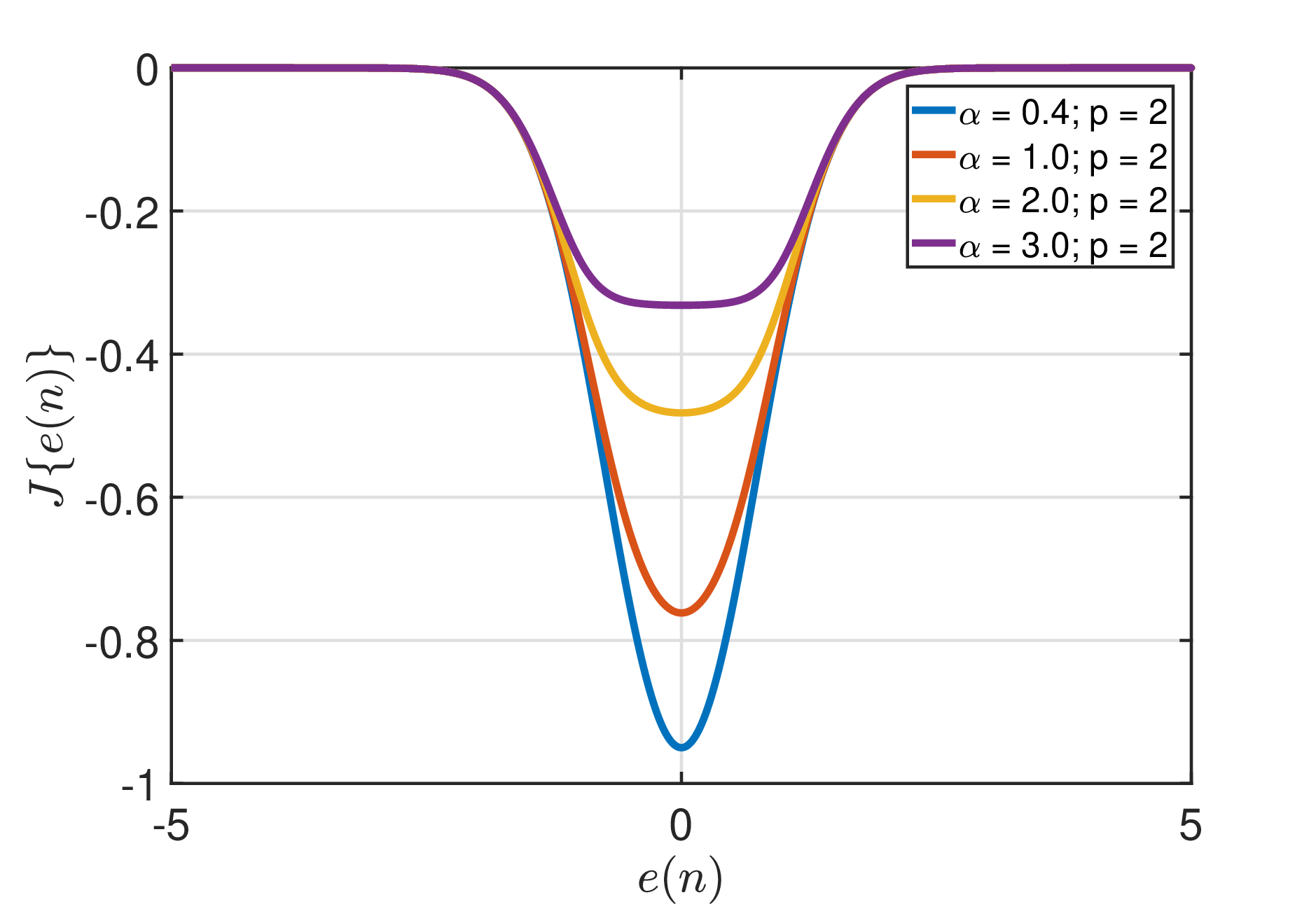}
\label{objective function alpha}}
\hfill
\subfloat[]
{\includegraphics[width=0.475\linewidth]{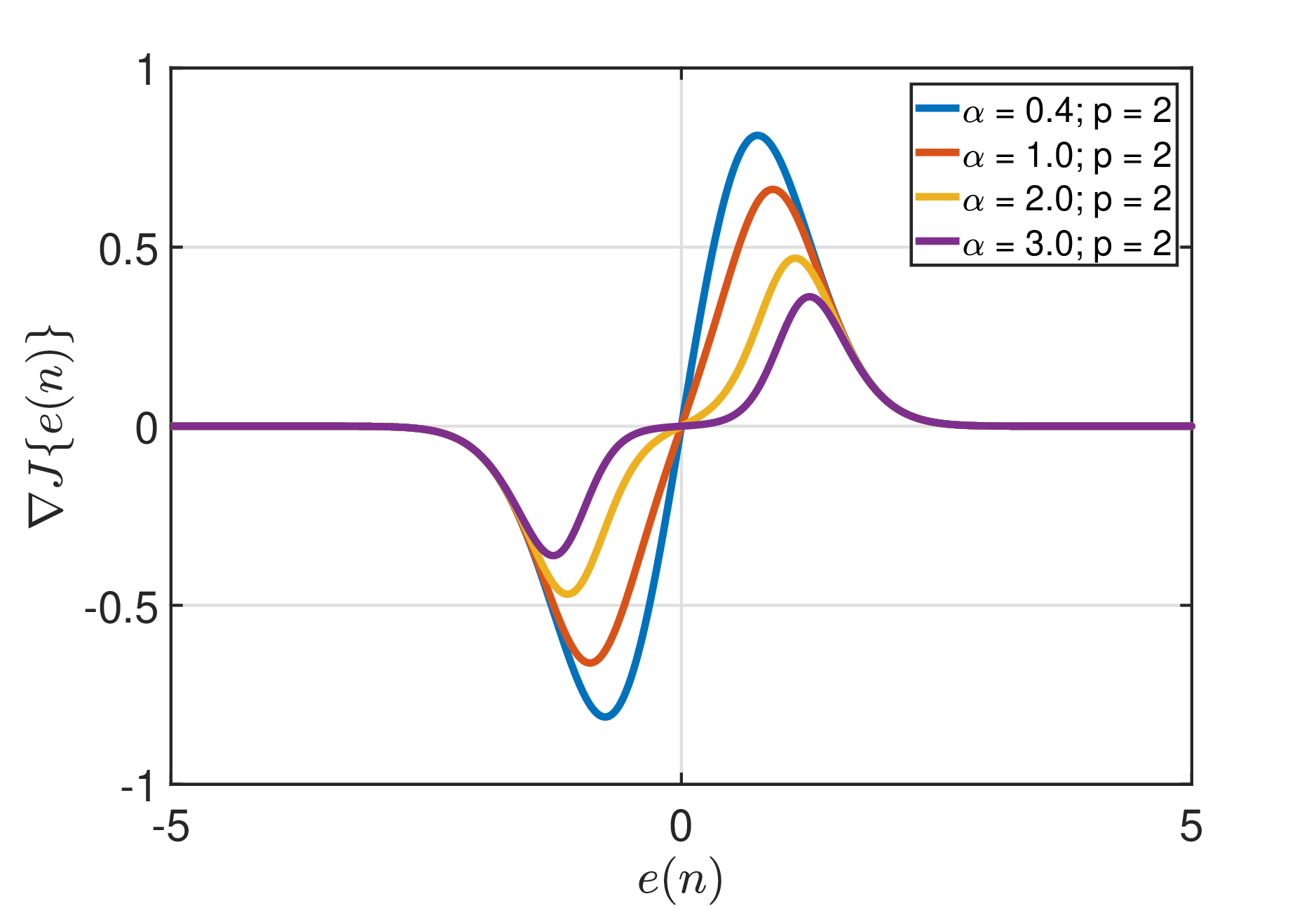}}

\label{objective function alpha derivative}
\vspace{-0.35em} 
\caption{\small Effect of $\alpha$ over {(a)} the objective function and {(b)} its derivative.}
\label{objt func alpha}
\end{figure*}

After some theoretical study and an experimental procedure for performance evaluation, the proposed FXHEKM approach employs an objective function based on the hyperbolic tangent of a kernel of the residual error computed in  \eqref{e}, which is defined by
\begin{equation}
\begin{aligned}
    \centering
    J\{\mathbf{w}(n)\} = -\rho \frac{1}{\alpha} \tanh \left(\alpha \exp^{-\eta|e(n)|^{p}}\right), 
    \label{Jw}
\end{aligned}
\end{equation}
where $\alpha$ $>$ 0 is a conveniently chosen constant, $\eta$, $p$ $>$ 0 are constants to set the exponential behavior of the kernel, {and the constant $\rho>0$ helps to set the learning rate.} Note that these parameters directly influence the algorithm's performance. From Fig. \ref{objective function p}. we can see that {larger values of $p$ yield larger values in the gradient, which affects the learning rate of the algorithm. However, extremely large values of $p$ result in flat segments in the derivative of the objective function (in the proximity of $e(n)=0$), which is detrimental to the filter operation.} Similarly, from Fig. \ref{objt func alpha}, we can see that smaller values of $\alpha$ lead to better performance.   

Taking the derivative of the cost function and using the principle of the gradient descent method, we obtain the expression for the term of the update:
\begin{equation}
\begin{aligned}
    \centering
    \dfrac{\partial J\{\mathbf{w}(n)\}}{\partial \mathbf{w}(n)} &= -\rho \ \mathrm{sech}^{2} \left(\alpha \exp^{-\eta|e(n)|^{p}}\right) \exp^{-\eta|e(n)|^{p}} \\ 
    & \quad \ \eta p |e(n)|^{p-1} \mathrm{sign}(e(n)) \mathbf{x}'(n) . 
    \label{dJdw}
\end{aligned}
\end{equation}

Thus, the expression of this partial derivative of the cost function is used to obtain the {update equation of the proposed FXHEKM algorithm as}
\begin{equation}
\begin{aligned}
\label{eq7}
    \centering
    \mathbf{w}(n+1) = \ &\mathbf{w}(n) - \nabla J\{\mathbf{w}(n)\} \\
    &\mathbf{w}(n) + {\mu \phi(e(n))} \mathbf{x}'(n),
\end{aligned}
\end{equation}
where 
{$\mu$ = $\rho$$\eta$\textit{p}} represents the value of the step-size and the function
\begin{equation}
\begin{aligned}
\label{eq7_1}
\phi(e(n)) = \ &\mathrm{sech}^{2} \left(\alpha \exp^{-\eta|e(n)|^{p}}\right) \exp^{-\eta|e(n)|^{p}} \\
&|e(n)|^{p-1} \frac{\mathrm{sign}(e(n))}{\delta + \lVert \mathbf{x}'(n) \lVert^{2}_{2} } ,
\end{aligned}
\end{equation}
{is introduced to simplify the notation. The term}
$\lVert \mathbf{x}'(n) \lVert^{2}_{2} \ > 0$ is a \textit{l}$^{2}$-norm of the input vector used for the final improved normalized version. As a result, an extremely small positive value $\delta$ is increased as a regularization factor to avoid division by zero.

Lastly, to obtain a reliable response concerning the learning curves variations, we introduce a robust strategy against non-Gaussian noises (\textit{i.e.} impulsive), by employing the function $q(e(n))$ based on the $M$-estimate method of the work of Yu \textit{et al} \cite{ref11} given by
\begin{equation}
\begin{aligned}
\label{eq7_2}
q(e(n)) = 
        \begin{cases}
            1, \ &|e(n)| < \zeta \\
            0, & |e(n)|\geq \zeta
        \end{cases},
\end{aligned}
\end{equation}
where $\zeta$ is the threshold coefficient of the $M$-estimate method that controls the response of this function in the update term in \eqref{eq7}, based on the comparison with the value of the modulus of the residual error.
{Therefore}, the definition for the update recursion of the weight vector $\mathbf{w}(n)$ employed by the proposed FXHEKM algorithm is given by
\begin{equation}
\begin{aligned}
\label{eq7_3}
    \centering
    \mathbf{w}(n+1) = \mathbf{w}(n) + \mu q(e(n))\phi(e(n)) \mathbf{x}'(n).
\end{aligned}
\end{equation}

As can be seen in what follows, in Algorithm \ref{alg1} the proposed FXHEKM algorithm is described and listed step-by-step in detail.

\begin{algorithm}
    \caption{FXHEKM Algorithm }\label{alg1}
    \begin{algorithmic}
        \REQUIRE $\mathbf{x}(n) = \left[ x(n) \ \ ... \ \ x(n - L + 1) \right]^{T}, 
        \newline \hspace{3.5cm}
        \mathbf{w}(n) = \left[ w_0(n) \ \ ... \ \ w_{L - 1}(n) \right]^{T}$
        \REQUIRE $d(n), P(z), S(z)$
        \REQUIRE $\mu, \ \zeta, \ \eta, \ \alpha, \ p, \ \rho$ and $\phi(e(n)) = 0$
        \WHILE {$n \leq$ {{$N$}}}
            \STATE $y(n) = \mathbf{x}(n)^T\mathbf{w}(n)$
            \STATE $e(n) = d(n) - s(n) * (\mathbf{w} (n)^{T}\mathbf{x}(n))$
            \IF {$e(n) < \zeta$}
                \STATE $\mathbf{w}(n+1) = \mathbf{w}(n) + \mu \phi(e(n)) \mathbf{x}'(n)$
            \ELSE
                \STATE $\mathbf{w}(n+1) = \mathbf{w}(n)$
            \ENDIF
        \ENDWHILE
    \end{algorithmic}
\end{algorithm}

\subsection{Computational complexity}

Using the recursive equation that describes the update rule of the weight vector, we can obtain the computational complexity required to compute the algorithms \cite{ref13}. Table \ref{tab1} shows the estimation of the computational cost in terms of multiplications/divisions, sums, and nonlinear operations - i.e. log(.), exp(.), sech(.) of the proposed FXHEKM and competing algorithms available in the literature \cite{ref8,ref9,ref10,ref11} and used as a comparison in this work.

\begin{table}[htbp]
\caption{Computation complexity of the proposed FXHEKM and competing algorithms per iteration}
\label{tab1}
\begin{center}
\begin{tabular}{|c|c|c|c|}
\hline
\hline
\textbf{Algorithm} & \textbf{
Mult} , \textbf{Div} & \textbf{+} & \textbf{Nonlinear Op} \\
\hline
FXLMS & 2L+1 & 2L+2M-3 & - \\
FXGMCC & 2L+4 & 2L+2M-3 & 3 \\
FXGHT & (p+5)(L-1) + L & 3L-2 & 1 \\ 
IFXGMCC & 2L+6 & 2L+2M-3 & 4 \\
FXGR & 2L, 1 & 2L & - \\
FXECHF & 2L+p+8, 1 & 2L+2 & 2 \\
FXHEKM & 3L+2p+7, 1 & 3L-2 & 4 \\
\hline
\hline
\end{tabular}
\end{center}
\end{table}
It should be noted that the proposed FHEKM has a computational cost that is higher than the standard FXLMS algorithm but comparable to recently reported algorithms such as FXECHF and IFXGMCC.

\section{Statistical Analysis}

This section describes a statistical analysis of the FXHEKM algorithm. In particular, we consider the stability conditions and the MSE at steady state for the proposed FXHEKM algorithm. This part of the study aims to ensure the basis that provides mathematical guarantees to the proposed approach as evidenced in the literature \cite{yang}. In particular, we obtain the stability conditions and derive formulas to predict the MSE of the proposed FXHEKM algorithm at steady state. To this end, we use common assumptions such as independence theory to obtain the stability, stationary, and other critical aspects of the proposed HEKM algorithm.

\subsection{Stability}

Let us begin the analysis with the update equation \eqref{eq7}, where we subtract the optimal weights $\mathbf{w_o}$ from both sides of the equation and define the error of the filter weights concerning the optimal weight vector as 
{
\begin{equation}
\label{eq3_6}
    \centering
    {\varepsilon}(n)=\mathbf{w_o}-\mathbf{w}(n).
\end{equation}}

Then, substituting the last Equation in (\ref{eq7_3}), we obtain
\begin{equation}
\begin{aligned}
\label{eq3_7}
    \centering
    {\varepsilon}(n+1) &= {\varepsilon}(n) {-} \mu \phi(e(n)) \mathbf{x}'(n) \\
    &= {\varepsilon}(n) - g(e(n)).
\end{aligned}
\end{equation}

As we can observe, the term related to the hyperbolic secant of an exponential can be difficult to manipulate algebraically to obtain an analytical solution. Thus, we consider as a boundary approximation the case of $p=2$  (standard form of FXHEKM) \cite{ref9}. 
{Note also that we can define from the Equations \eqref{e} and \eqref{eq3_6}, a useful error expression, which will be widely used and is described by}
\begin{equation}
\begin{split}
\label{eq3_72}
    e(n) &= d(n) - \mathbf{w}^{\text{H}}_{\text{o}}\mathbf{x}'(n) + {\varepsilon}(n)\mathbf{x}'(n) \\
    & \quad \ e_{\text{o}}(n) + {\varepsilon}(n)\mathbf{x}'(n) ,
\end{split}
\end{equation}
{where $e_{\text{o}}(n)=d(n) - \mathbf{w}^{\text{H}}_{\text{o}}\mathbf{x}'(n)$ is the optimal error value.}

Taking into account the last equation and the term referred to as $g(e(n))$ in \eqref{eq3_7}, which is a function-related  derivative, where we decouple the error to manipulate some steps that permit the stability analysis, we obtain
\begin{equation}
\begin{split}
\label{eq3_8}
    g(e(n)) &= \mu \phi(e(n)) \\
    & \quad \ |d(n) -\mathbf{x}'(n)^{\text{H}}\mathbf{w}(n)| \mathbf{x}'(n) \hspace{0.15 cm} \\
    &= \mu \phi(e(n)) \\ 
    & \quad \ |e^{*}_{\text{o}}(n) + \mathbf{w}^{\text{H}}_{\text{o}}\mathbf{x}'(n) - \mathbf{x}'(n)^{\text{H}}\mathbf{w}(n)| \mathbf{x}'(n) \\
    &= \mu \phi(e(n)) \\ 
    & \quad \ (\mathbf{x}'(n)e^{*}_{\text{o}}(n) + (\mathbf{x}'(n)^{\text{H}}\mathbf{x}'(n) {\varepsilon}(n))).
\end{split}
\end{equation}
Note that we consider here $\delta$ = 0 for the sake of simplicity. Returning to equation \eqref{eq3_7}, applying the expected value operator, and rearranging the terms, we have
\begin{equation}
\begin{split}
\label{eq3_9}
    \centering
    E\left[ {\varepsilon}(n+1)\right] &= E\left[\left( {\mathrm{I}} - \left(\mu \phi(e(n))\right) \mathbf{x}'(n)^{\mathrm{H}}\mathbf{x}'(n)\right) {\varepsilon}(n)\right] \\
    & - E\left[\left(\mu \phi(e(n))\right) \mathbf{x}'(n)e_{\text{o}}(n)\right].
\end{split}
\end{equation}

Now, we assume that \textbf{x}'(n) is independent from $ {\varepsilon}(n)$ and $e_{\text{o}}(n)$. Using this assumption {and the eigenvalue decomposition of the covariance matrix, i.e.,  $\mathbf{R_{x'}} =
\mathbf{V} {\Lambda}\mathbf{V}^{\mathrm{H}}$ where $\mathbf{V}$ is a unitary matrix, \textit{i.e.} $\mathbf{V}^{\mathrm{H}}\mathbf{V}$ = $\mathbf{V}^{-1}\mathbf{V}$ = $\mathbf{I}$ and $\mathbf{\Lambda}$ is a diagonal matrix with the eigenvalues of  $\mathbf{R_{x'}}$}, we arrive at
{
\begin{equation}
\begin{aligned}
\label{eq3_10}
    \centering
    E\left[ {\varepsilon}(n+1)\right] &
    = \left( {\mathbf{I}} - \mu E\left[\phi(e(n)) \right] \mathbf{V} {\Lambda}\mathbf{V}^{\mathrm{H}}\right) E\left[ {\varepsilon}(n)\right].
\end{aligned}
\end{equation}
}
{The expression in \eqref{eq3_10} is the nonlinear stochastic difference equation for the stability condition of the FXHEKM algorithm.} 

Multiplying both sides by the unitary matrix $\mathbf{V}$ and rearranging the terms, we get
\begin{equation}
\begin{aligned}
\label{eq3_13}
    \centering
    E\left[ {\varepsilon}'(n+1)\right] = &\left( {\mathrm{I}} - \mu E\left[ \phi(e(n)) \right] {\Lambda}\right) E\left[ {\varepsilon}'(n)\right] .
\end{aligned}
\end{equation}

For the evaluation of the convergence of the recursion of the FXHEKM algorithm, we can rewrite the recursion step as
\begin{equation}
\begin{aligned}
\label{eq3_14}
    \centering
    E\left[ {\varepsilon}'(3)\right] &= \left( {\mathrm{I}} - \mu E\left[\phi (e(2)) \right] {\Lambda}\right) E\left[ {\varepsilon}'(2)\right] \\
    &= \left( {\mathrm{I}} - \mu E\left[ \phi(e(1)) \right] {\Lambda}\right)^{2} E\left[ {\varepsilon}'(1)\right] .
\end{aligned}
\end{equation}

Using this recursion to {$n = l$} elements, we obtain 
\begin{equation}
\begin{aligned}
\label{eq3_15}
    \centering
    E\left[ {\varepsilon}'(l+1)\right] = &\left( {\mathrm{I}} - \mu E\left[ \phi(e(l)) \right] {\Lambda}\right)^{l} E\left[ {\varepsilon}'(1)\right] ,
\end{aligned}
\end{equation}
and taking into account the need to use only the diagonal elements, we can apply the decoupling property in the error weight vector. Therefore, we can obtain a simplified form for the \textit{k}-th component of $ {\varepsilon}'(n)$ given by 
\begin{equation}
\begin{aligned}
\label{eq3_16}
    \centering
    E\left[ {\varepsilon}'_{k}(n+1)\right] = &\left(1 - \mu E\left[\phi(e(n)) \right] {\lambda}_{k}\right)^{n+1}E\left[ {\varepsilon}'_{k}(0)\right] ,
\end{aligned}
\end{equation}
where the expression above is function of the singular value $ {\varepsilon}'_{k}(0)$, and the $ {\lambda}_{k}$ is the \textit{k}-th component of $ {\Lambda}$. Then, analyzing the condition that guarantees convergence of the FXHEKM algorithm, the coefficients in the mean, we need to satisfy that
\begin{equation}
\begin{aligned}
\label{eq3_17}
    \centering
    \left|1 - \mu E\left[ \phi(e(n)) \right] {\lambda}_{k}\right| < 1 \hspace{0.15 cm} ,
\end{aligned}
\end{equation}
where the eigenvector {$ \lambda_{k}$} is the set of eigenvalues related to the autocorrelation matrix $\mathbf{R_{x}}$ that solves the Wiener filter problem. Expanding the expression and isolating the term of the step-size, we arrive at an expression for the stability condition given by
\begin{equation}
\begin{aligned}
\label{eq3_18}
    \centering
    &-1 < 1 - \mu E\left[\phi(e(n)) \right] {\lambda}_{k} < 1 \ ,
\end{aligned}
\end{equation}

and finally, we obtain that

\begin{equation}
\begin{aligned}
\label{eq3_20}
    \centering
    0 < \mu < \frac{2}{ {\lambda}_{max}  {\Phi}(e(n))} \ ,
\end{aligned}
\end{equation}
where {$ {\Phi}(e(n)) = \ E\left[ \phi(e(n))\right]$. Expanding $ {\Phi}(e(n))$, we get}
{\begin{align}
     {\Phi}&(e(n)) \nonumber\\
    &\ =E\left[\text{sech}^{2}\left(\alpha \exp^{\eta|e(n)|^{p}}\right) \exp^{\eta|e(n)|^{p}} \frac{\mathrm{sign}(e(n))}{\delta + \lVert \mathbf{x}'(n) \lVert^{2}_{2}}\right] \nonumber\\
    &\approx \ \frac{1}{N}\sum_{n=0}^{N-1}\left(\phi(e(n))\right),\label{eq3_22}
\end{align}}which is the analytical expression that represents a function of the residual error defined by the expected value of a resulting component function of the FXHEKM algorithm, which is calculated in the learning process.


Lastly, we still have another alternative expression for the stability condition. In the case of $p = 2$ (a special case of FXHEKM), we have
\begin{equation}
\begin{aligned}
\label{eq3_23}
    0 < \mu < \frac{1}{ {\lambda}_{max}  {\Phi}(e(n))},
\end{aligned}
\end{equation}
{where equations \eqref{eq3_18}, \eqref{eq3_20} and \eqref{eq3_23}} represent different formulations to express the condition of the stability of the proposed FXHEKM algorithm. 

\subsection{Steady-state MSE}
\label{ss_mse}
In this section, we analyze the MSE of the proposed FXHEKM algorithm when the steady state is reached. Firstly, we employ the MSE function that can be written as
\begin{equation}
\begin{aligned}
\label{eq3_25}
    \centering
    J(n) &= E\left[ |e(n)|^{2} \right] \\
    &= E\left[(e_{\text{o}}(n) +  {\varepsilon}^{\mathrm{H}}(n) \mathbf{x}'(n))^{*} (e_{\text{o}}(n) +  {\varepsilon}^{\mathrm{H}}(n) \mathbf{x}'(n))\right] \\
    &= E\left[e_{\text{o}}(n)^{*}e_{\text{o}}(n)\right] + E\left[e_{\text{o}}(n)^{*} {\varepsilon}^{\mathrm{H}}(n) \mathbf{x}'(n)\right] \\
    &+ E\left[e_{\text{o}}(n)\mathbf{x}'^{\mathrm{H}}(n) {\varepsilon}(n)\right] + E\left[\mathbf{x}'^{\mathrm{H}}(n) {\varepsilon}(n) {\varepsilon}^{\mathrm{H}}(n) \mathbf{x}'(n)\right] .
\end{aligned}
\end{equation}
As we know, we use the concept of statistical independence from $e_{\text{o}}(n)$ to $\varepsilon(n)$ and $\mathbf{x}'$ \cite{ref2}. Then, we obtain
\begin{equation}
\begin{aligned}
\label{eq3_26}
    \centering
    J(\infty) &= J_{\mathrm{min}} + J_{ex} \\
    &= J_{\mathrm{min}} + \mathbf{tr}\{\mathbf{K}(n)\mathbf{R}_{x}'\} \\
    &= J_{\mathrm{min}} + \mathbf{tr}\{\mathbf{\Lambda}\mathbf{U}\} \\
    &= J_{\mathrm{min}} + \sum_{i=1}^{M} \lambda_i u_{i,i}(n),
\end{aligned}
\end{equation}
is the function-based calculation of the excess value of the MSE estimation. Using the equation for the error vector in \eqref{eq3_10}, we get
\begin{equation}
\begin{split}
\label{eq3_27}
    \centering
    J(\infty) =& \ J_{\mathrm{min}} + \mathbf{tr}\left\{\left( {\mathrm{I}} - \mu E\left[ \phi(e(n)) \right]
   \mathbf{R}_{x}'\right) \right. \\ 
   & \ \ \ \mathbf{K}(n-1)\left. \left( {\mathrm{I}} - \mu E\left[ \phi(e(n)) \right]
   \mathbf{R}_{x}'\right)\right\} ,
\end{split}
\end{equation}
where the second term on the right side 
is the excess value of the MSE estimation {and $\phi(e(n))$ was defined in \eqref{eq7_1}}. An important step here is the calculation of the covariance matrix of the weight error vector. From the definition of ${\mathbf K}(n)$ and using \eqref{eq3_13}, the desired expression for ${\mathbf K}(n)$ is given by 
\begin{equation}
\begin{aligned}
\label{eq3_29}
    \centering
    \mathbf{K}(n+1) = E\left[  {\varepsilon}(n+1)  {\varepsilon}^{\mathrm{H}}(n+1)\right],
\end{aligned}
\end{equation}
where by extending the expression of the weight error vector, we have
\begin{equation}
\begin{aligned}
\label{eq3_30}
    \centering
    &\mathbf{K}(n+1) = \\
    & E\left[\left( {\mathrm{I}} - \left(\mu \phi(e(n))\right) \mathbf{x}'(n)^{\mathrm{H}}\mathbf{x}'(n)\right)  {\varepsilon}(n) \right. \\
    &\ \ \ \ \ \ \ \ \ \left. - \left(\mu \phi(e(n))\right) \mathbf{x}'(n)e_{\text{o}}(n) \right] = \\
    & E\left[\left( {\mathrm{I}} - \left(\mu \phi(e(n))\right) \mathbf{x}'(n)^{\mathrm{H}}\mathbf{x}'(n)\right)  {\varepsilon}(n) \right. \\
    &\ \ \ \ \ \ \ \ \ \left . - \left(\mu \phi(e(n))\right) \mathbf{x}'(n)e_{\text{o}}(n) \right]^{\mathrm{H}} \\
\end{aligned}
\end{equation} 
Organizing the terms, we get 
\begin{equation}
\begin{aligned}
\label{eq3_32}
    \centering
    \mathbf{K}&(n+1) = \\
    & \ \ \ \ \  E\left[\left( {\mathrm{I}} - \left(\mu\phi(e(n))\right) \mathbf{x}'(n)^{\mathrm{H}}\mathbf{x}'(n)\right)  {\varepsilon}(n)  {\varepsilon}(n)^{\mathrm{H}} \right. \\
    & \quad \ \ \ \ \ \ \left. \left( {\mathrm{I}} - \left(\mu \phi(e(n))\right)^{*}\mathbf{x}'(n)\mathbf{x}'(n)^{\mathrm{H}}\right) \right] \\
    & - E\left[\left( {\mathrm{I}} - \left(\mu \phi(e(n))\right) \mathbf{x}'(n)^{\mathrm{H}}\mathbf{x}'(n)\right)  {\varepsilon}(n) \right. \\
    & \qquad \ \ \left. \left(\mu \phi(e(n))\right) \mathbf{x}'(n)^{\mathrm{H}}e_{\text{o}}(n) \right] \\
    & - E\left[ \left( {\mathrm{I}} - \left(\mu \phi(e(n))\right)^{*} \mathbf{x}'(n)\mathbf{x}'(n)^{\mathrm{H}}\right)    {\varepsilon}(n)^{\mathrm{H}}  \right. \\
    & \qquad \ \ \left. \left(\mu \phi(e(n))\right) \mathbf{x}'(n)^{\mathrm{H}}e_{\text{o}}(n) \right] \\
    & + E\left[ \mu^{2} |\phi(e(n))|^{2} |e_{\text{o}}(n)|^{2} \mathbf{x}'(n)\mathbf{x}'(n)^{\mathrm{H}}\right] .
\end{aligned}
\end{equation}

Using the assumption of statistical independence for $e_{\text{o}}(n)$ and $ {\varepsilon}(n)$, and assuming the orthogonality principle between $e_{\text{o}}(n)$ and $\mathbf{x}'(n)$, we arrive at the last equation divided in $\textbf{T}_{i}$ terms:
\begin{equation}
\begin{aligned}
\label{eq3_33}
    \centering
    &\mathbf{T}_{1} = \\
    &E \left[ ( {\varepsilon}(n)  {\varepsilon}(n)^{\mathrm{H}} - \mu \phi(e(n)) \mathbf{x}'(n)\mathbf{x}'(n)^{\mathrm{H}}  {\varepsilon}(n)  {\varepsilon}(n)^{\mathrm{H}}) \right. \\ 
    & \qquad\ \quad \ 
    \left. ( {\mathrm{I}} - \mu \phi(e(n))^{*}\mathbf{x}'(n)\mathbf{x}'(n)^{\mathrm{H}}) \right] \\
    &\quad \ = \mathbf{K}(n) - \mu E [\phi(e(n))] \mathbf{R}_{x}' \mathbf{K}(n) \\
    & \quad \quad \ \ \ \ \ \ - \mu E [ \phi(e(n)) ] \mathbf{K}(n) \mathbf{R}_{x}' + \mu^{2} \mathbf{R}_{x}' \mathbf{K}(n) \mathbf{R}_{x}' \\
    &\quad \ = ( {\mathrm{I}} - \mu \mathbf{R}_{x}' \Phi(e(n))) \mathbf{K}(n) ( {\mathrm{I}} - \mu \mathbf{R}_{x}' \Phi(e(n))^{*}) \ ;
\end{aligned}
\end{equation}

\begin{equation}
\begin{aligned}
\label{eq3_34}
    \centering
    \mathbf{T}_{2} = \mathbf{T}_{3} = 0 \ ;
\end{aligned}
\end{equation}

\begin{equation}
\begin{aligned}
\label{eq3_35}
    \centering
    \mathbf{T}_{4} &= E \biggl[ \mu^{2} |\phi(e(n))|^{2} |e_{\text{o}}(n)|^{2}\mathbf{x}'(n)\mathbf{x}'(n)^{\mathrm{H}} \biggr]  \\
    &= \mu^{2} |\Phi(e(n))^{2}|J_{min} \mathbf{R}_{x}' \ .
\end{aligned}
\end{equation}
Eventually, we obtain the following expression for the covariance matrix:
\begin{equation}
\begin{aligned}
\label{eq3_36}
    \centering
    &\mathbf{K}(n+1) = \\
    &( {\mathrm{I}} - \mu \Phi(e(n)) \mathbf{R}_{x}') \mathbf{K}(n)( {\mathrm{I}} - \mu \Phi(e(n)) \mathbf{R}_{x}') \\
    & \qquad \qquad + \mu^{2} |\Phi(e(n))^{2}|J_{min} \mathbf{R}_{x}' \ .
\end{aligned}
\end{equation}
Multiplying both sides by the unitary matrices $\mathbf{V}$ and $\mathbf{V}^{\mathrm{H}}$, we get
\begin{equation}
\begin{aligned}
\label{eq3_37}
    \centering
    &\mathbf{V}^{\mathrm{H}}\mathbf{K}(n+1)\mathbf{V} = \\
    &\mathbf{V}^{\mathrm{H}} ( {\mathrm{I}} - \mu \Phi(e(n)) \mathbf{R}_{x}') \mathbf{K}(n) ( {\mathrm{I}} - \mu \Phi(e(n)) \mathbf{R}_{x}')\mathbf{V} \\
    & \qquad \qquad \quad \ + \mu^{2} |\Phi(e(n))^{2}|J_{min} \mathbf{V}^{\mathrm{H}}\mathbf{R}_{x}'\mathbf{V} \ .
\end{aligned}
\end{equation}
Then, simplifying the terms, we reach
\begin{equation}
\begin{aligned}
\label{eq3_38}
    \centering
    &\mathbf{U}(n+1) = \\
    &(\mathbf{V}^{\mathrm{H}} - \mu \Phi(e(n)) \mathbf{\Lambda}\mathbf{V}^{\mathrm{H}} ) \mathbf{K}(n) (\mathbf{V} - \mu \Phi(e(n)) \mathbf{\Lambda}\mathbf{V}) \\
    & \qquad \qquad \quad \ + \mu^{2} |\Phi(e(n))^{2}|J_{min} \mathbf{\Lambda} \ .
\end{aligned}
\end{equation}
Therefore, the matrix $\mathbf{U}(n+1)$ can be computed recursively as 
\begin{equation}
\begin{aligned}
\label{eq3_39}
    \centering
    \mathbf{U}(n+1) &= ( {\mathrm{I}} - \mu \Phi(e(n)) \mathbf{\Lambda}) \mathbf{U}(n)( {\mathrm{I}} - \mu \Phi(e(n)) \mathbf{\Lambda}) \\
    & \quad \ + \mu^{2} |\Phi(e(n))^{2}|J_{min} \mathbf{\Lambda} \ .
\end{aligned}
\end{equation}

To calculate the excess MSE we need the diagonal elements of the matrices, which are described by
\begin{equation}
\begin{aligned}
\label{eq3_40}
    \centering
    \mathbf{u}_{i,i}(n+1) &= (1 - \mu^{2} \Phi(e(n))^{2} \mathbf{\lambda}_i)^{2} \mathbf{u}_{i,i}(n) \\
    &+ \mu^{2} |\Phi(e(n))^{2}|J_{min} \mathbf{\lambda}_i \ .
\end{aligned}
\end{equation}
Based on \eqref{eq3_40} and obtaining the expression of \eqref{eq3_26} in the steady state, we consider the performance of the FXHEKM algorithm at steady state. The diagonal elements are given by
\begin{equation}
\begin{split}
\label{eq3_41}
    \centering
    \mathbf{u}_{i,i}(\infty) &= (1 - \mu^{2} \Phi(e(n))^{2} \mathbf{\lambda}_i)^{2} \mathbf{u}_{i,i}(\infty) \\
    &\quad \ + \mu^{2} |\Phi(e(n))^{2}|J_{min} \mathbf{\lambda}_i \ \\
    &= \mathbf{u}_{i,i}(\infty) - 2 \mu^{2} \Phi(e(n))^{2} \mathbf{\lambda}_i \mathbf{u}_{i,i}(\infty) \\ 
    &\quad \ + \mu^{4} \Phi(e(n))^{4} \mathbf{\lambda}_i^{2}\mathbf{u}_{i,i}(\infty) \\
    &\quad \ + \mu^{2} |\Phi(e(n))^{2}|J_{min} \mathbf{\lambda}_i \ .
\end{split}
\end{equation}
Manipulating the products of the above equation, arranging and simplifying the terms of the previous expression, we have
\begin{equation}
\begin{split}
\label{eq3_42}
    \centering
    2 \mu^{2} \Phi(e(n))^{2} &\mathbf{\lambda}_i \mathbf{u}_{i,i}(\infty) - \mu^{4} \Phi(e(n))^{4} \mathbf{\lambda}_i^{2}\mathbf{u}_{i,i}(\infty) \\
    &= \mu^{2} |\Phi(e(n))^{2}|J_{min} \mathbf{\lambda}_i \ ,
\end{split}
\end{equation}
\begin{equation}
\begin{aligned}
\label{eq3_43}
    \centering
    2\mathbf{u}_{i,i}(\infty) - \mu^{2} \Phi(e(n))^{2} \mathbf{\lambda}_i\mathbf{u}_{i,i}(\infty) = J_{min} \ ,
\end{aligned}
\end{equation}

\begin{equation}
\begin{aligned}
\label{eq3_44}
    \centering
    \mathbf{u}_{i,i}(\infty)\left(2 - \mu^{2} \Phi(e(n))^{2}\mathbf{\lambda}_i\right) = J_{min} \ .
\end{aligned}
\end{equation}
Therefore, we obtain
\begin{equation}
\begin{aligned}
\label{eq3_45}
    \centering
    \mathbf{u}_{i,i}(\infty) = \frac{J_{min}}{2 - \mu^{2} \Phi(e(n))^{2}\mathbf{\lambda}_i} \ .
\end{aligned}
\end{equation}
Thus, the excess MSE of the FXHEKM algorithm at steady-state is given by 

\begin{equation}
\begin{aligned}
\label{eq3_46}
    \centering
    J_{ex}(\infty) = J_{min} \sum_{i = 1}^{M}\frac{\mathbf{\lambda}_i}{2 - \mu^{2} \Phi(e(n))^{2}\mathbf{\lambda}_i} \ .
\end{aligned}
\end{equation}

At last, it is interesting to point out the relation of the misalignment of the adjustment factor, which can be expressed as
\begin{equation}
\begin{aligned}
\label{eq3_47}
    \centering
   \mathcal{M} = \sum_{i = 1}^{M}\frac{\mathbf{\lambda}_i}{2 - \mu^{2} \Phi(e(n))^{2}\mathbf{\lambda}_i} \ ,
\end{aligned}
\end{equation}
and then by returning to \eqref{eq3_26}, assuming that the algorithm will be in the steady state and applying the expression obtained in \eqref{eq3_44}-\eqref{eq3_45}, we have

\begin{equation}
\begin{aligned}
\label{eq3_48}
    \centering
    J(\infty) &= J_{\mathrm{min}} \left( 1 + \mathcal{M} \right) \\
    &= J_{\mathrm{min}} \left( 1 + \sum_{i = 1}^{M}\frac{\mathbf{\lambda}_i}{2 - \mu^{2} \Phi(e(n))^{2}\mathbf{\lambda}_i} \right) ,
\end{aligned}
\end{equation}
where the formula of the steady-state MSE expression of the FXHEKM algorithm can be used to predict its MSE performance. Initially, we consider a standard simplification of
\begin{equation}
\begin{aligned}
    \label{eq3_49}
    \centering
    J_{\mathrm{min}} = \sigma^{2}_{v} \ ,
\end{aligned}
\end{equation}
as a reasonable value for the MMSE in the steady state.

\section{Simulations Results}
\label{sim_results}

In this section, the performance of the proposed FXHEKM algorithm is compared with standard and recent robust algorithms. The algorithms that employ a filtered-X framework are: the Least-Mean Square (FXLMS); the generalized Maximum Correntropy Criterion (FXGMCC) and its Improved version using the generalized Kernel (IFXGMCC) \cite{ref8}; the generalized Hyperbolic Tangent function - \textit{i.e.} GHT criterion (FXGHT) \cite{ref9} and an exponential hyperbolic cosine conjugated version (FXECH) \cite{ref10}; the M-estimator algorithm (FXGR) \cite{ref11} and the proposed FXHEKM algorithm. {The comparisons are restricted to the proposed FXHKEM algorithm with the aforementioned competing algorithms because they all consider linear ANC and constitute the most high-performance strategies for robust ANC. Other comparisons are possible but will be left for future work.}



In order to assess the robustness of the proposed FXHEKM algorithm, we consider noise with Gaussian and impulsive nature as input to the ANC system. Additive white Gaussian noise constitutes a standard noise in research, while impulsive noise is an interesting special case due to its low probability but high amplitude variation and broad spectral content. Based on the literature, we adopt an $\alpha$-stable distribution to generate both the additive white Gaussian noise and the impulsive noise. The characteristic function of the $\alpha$-stable distribution is given by 
\begin{equation}
\begin{aligned}
\label{eq15}
        \varpi(t) = \exp \{j\delta t - \gamma |t|^{\alpha_s}\left[ 1 + j\beta \text{sign}(t) S(t,\alpha_s)\right]\} \hspace{0.1 cm},
\end{aligned}
\end{equation}
where $\beta \in [-1,1]$ is the symmetry parameter, $\gamma>0$ is the dispersion parameter, $\delta \in \mathbb{R}$  is the location parameter, $\alpha_s$ denotes the characteristic factor of the distribution, and $S(t,\alpha_s)$ is given by

\begin{equation}
\begin{aligned}
\label{eq16}
        S(t,\alpha_s) = 
        \begin{cases}
            \frac{2}{\pi} \log|t|,& \text{if} \ \alpha_s = 1 \\
        \text{tan}(\frac{\alpha_s
        \pi}{2}),& \text{otherwise}
        \end{cases}.
\end{aligned}
\end{equation}
It is important to note that a smaller $\alpha_s$ yields a more impulsive noise.

The ANC system employed in this study included two scenarios: $\alpha$ = 2.0 (Gaussian noise) and 1.5 (pseudo-impulsive signal). The other three parameters are defined as standard $\alpha$-stable, i.e., $\beta$ = 0, $\gamma$ = 1 and $\delta$ = 0. The input noise $x(n)$ that follows an $\alpha$-stable distribution was multiplied by a factor equal to 0.1 (to avoid extrapolating peak values of scenario 2). Moreover, $v(n)$ is the independent measurement noise that follows a Gaussian distribution with zero mean and variance $\sigma_v^2$ that was set using white noise with 1\% of amplitude. To obtain the simulation results, we considered a filter length \textit{L} = 16, and 250 Monte Carlo trials.
The impulsive responses of the primary and secondary paths are given by \textit{P}(\textit{z}) = $0.25\textit{z}^{-2}+0.5\textit{z}^{-3}+1.0\textit{z}^{-4}+0.5\textit{z}^{-5}+0.25\textit{z}^{-6}$ and \textit{S}(\textit{z}) = 0.5\textit{P}(\textit{z}), respectively.

\begin{figure}[htbp]
\centerline{\includegraphics[scale=.275]{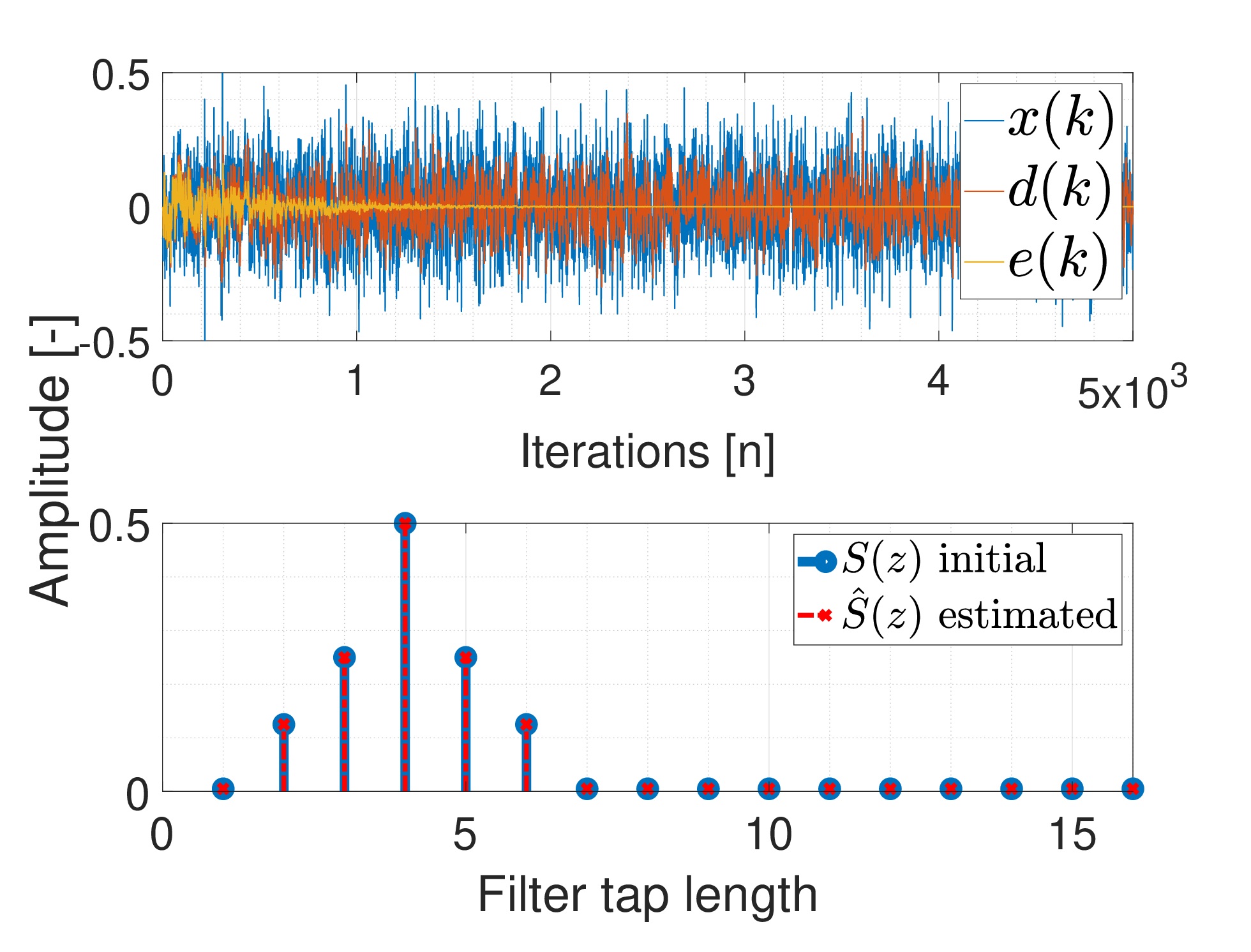}}
\caption{Initial system identification performed to estimate the secondary path $\hat{S}(z)$ in scenario 1.}
\label{fig2_1}
\end{figure}

Estimating the secondary path is a crucial step in filtered-X ANC (\textit{vide} \cite{ref16}). In this case history, an initial simulation is applied to define the initial values for the secondary path. Fig. \ref{fig2_1} shows that the estimation process of the secondary path is carried out effectively. This leads to accurate initial values, which are used to run the ANC system in the remaining simulations. In particular, we note in the figure that the estimation error (yellow line) decreases quickly, reaching a value close to zero, which demonstrates the effectiveness of the estimation process. We also note that the estimate $\hat{S}(z)$ matches the secondary path $S(z)$, as expected. 


\begin{figure}[ht!] 
    \begin{center}
        \includegraphics[scale=.26]{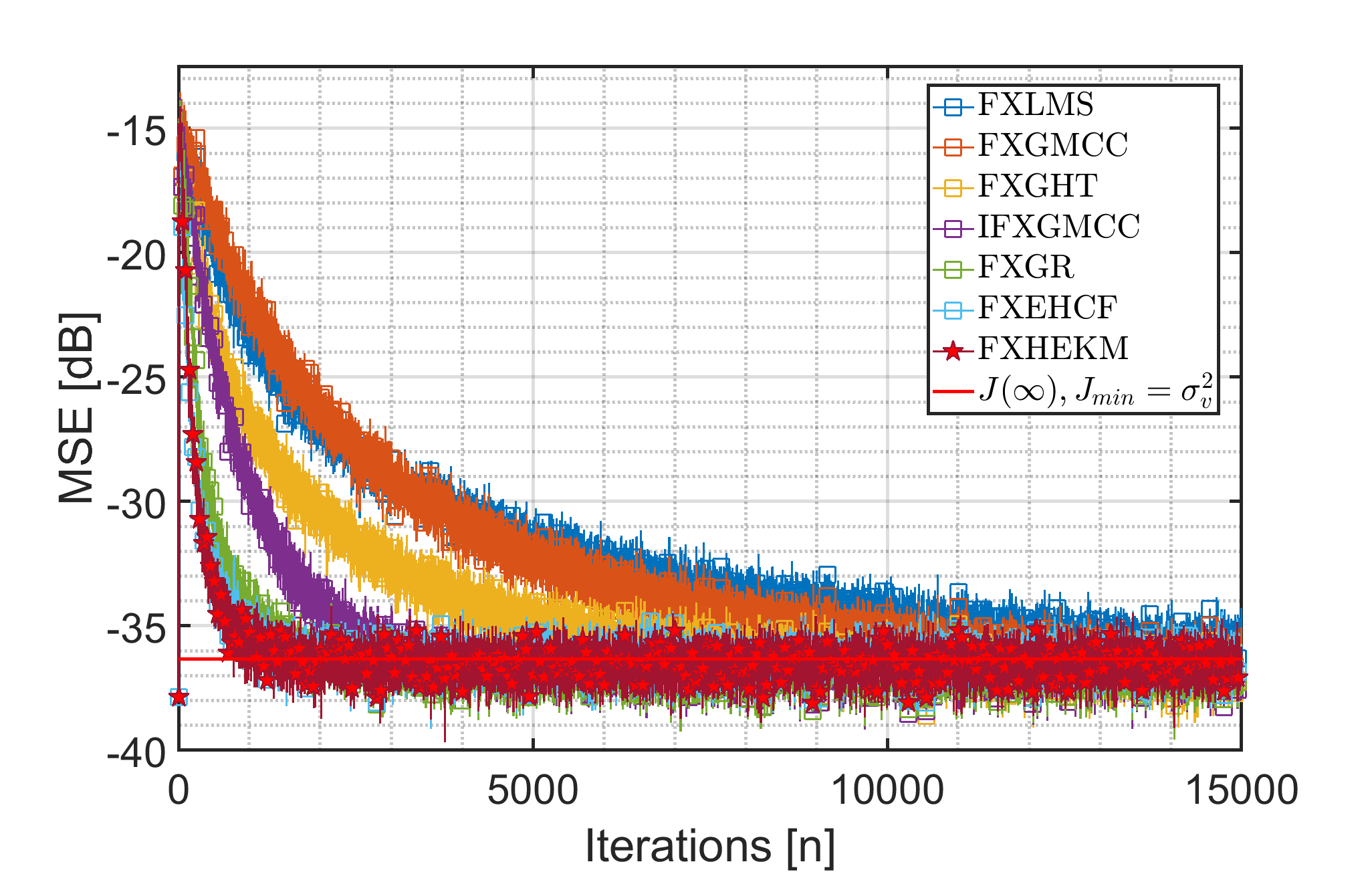}
        \caption{MSE performance comparison of the adaptive algorithms for Gaussian distribution noise (SNR = 20 dB).}  
        \label{MSE1_Mets_eta08_SNR20}
    \end{center}
\end{figure}

Next, we assess the MSE performance of the proposed FXHEKM and existing algorithms for an SNR equal to 20 dB, considering the best settings of the step-size and other parameters as depicted by Figure \ref{MSE1_Mets_eta08_SNR20}. We note that the proposed FXHEKM algorithm has a faster convergence than other robust and standard techniques, maintaining its stability over the iterations. {In addition, the simulated curves of FXHEKM agree well with the analytical result obtained in \eqref{eq3_46}, which is indicated by $J(\infty)$.} 

In the following examples, we compare the performance of different algorithms in terms of average noise reduction (ANR). The ANR performance metric is defined as follows \cite{ref8} $-$\cite{ref9}:
\begin{equation}
    \text{ANR}(n) = 20 \log \frac{A_{e(n)}}{A_{d(n)}} ,
\end{equation}
where $A_{e(n)} = \theta A_{e(n-1)} + (1-\theta)|e(n)|$, with $A_{e(0)} = 0$ denoting the estimate of residual error; $A_{d(n)} = \theta A_{d(n-1)} + (1-\theta)|d(n)|$, with $A_{d(0)} = 0$ describing the estimate of the noise in the primary path, and $\theta = 0.99$ is the forgetting factor.

\begin{figure}[ht!]
     \begin{center}
     \includegraphics[width=1.\columnwidth]{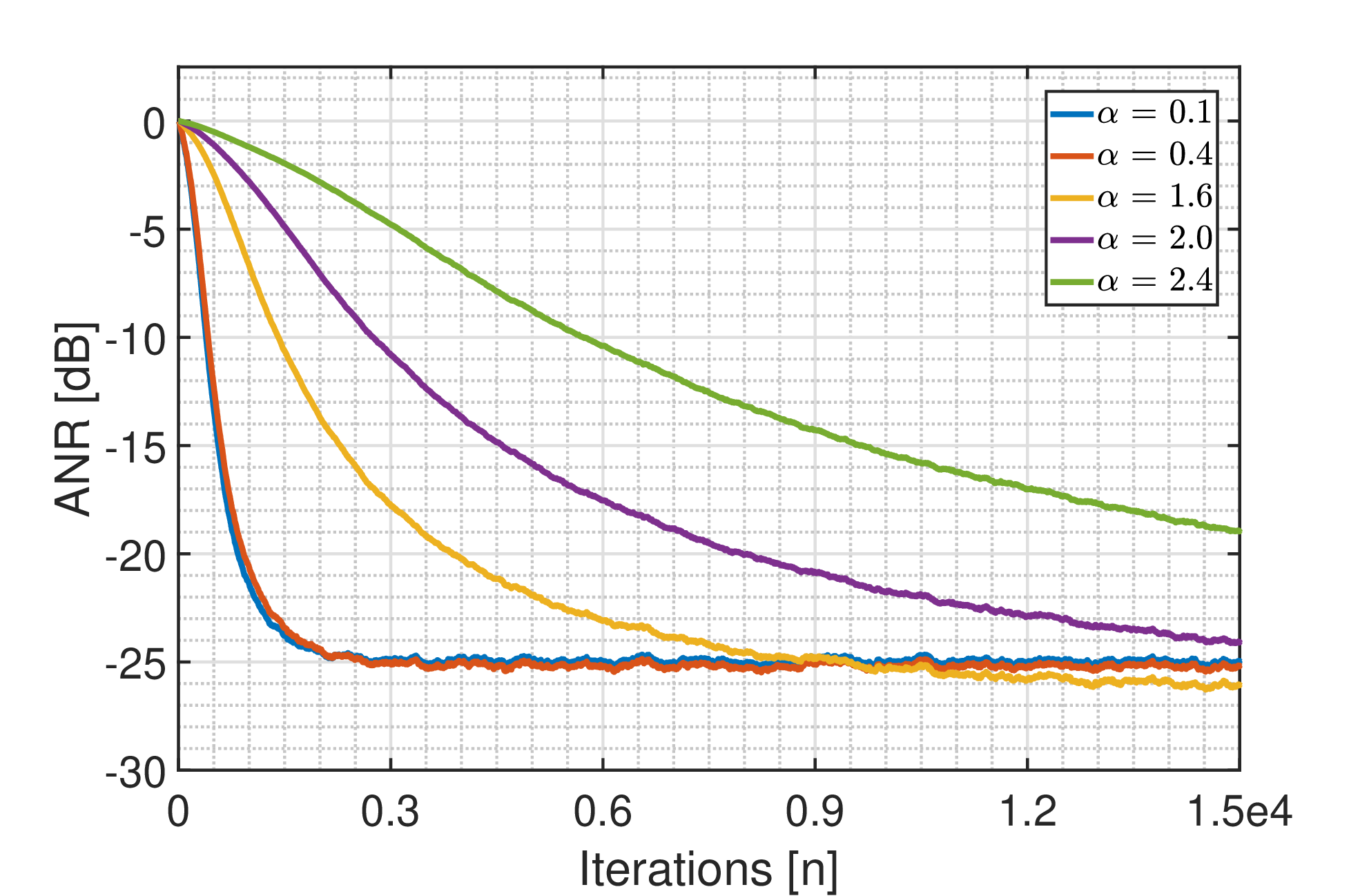}
         \caption{ANR performance at different values of algorithm's parameter $\alpha$ ($\alpha_s$ = 1.5).}
         \label{HEKM_alfa}
     \end{center}
 \end{figure}

\begin{figure}[ht!]
     \begin{center}
     	\includegraphics[width=1.\columnwidth]{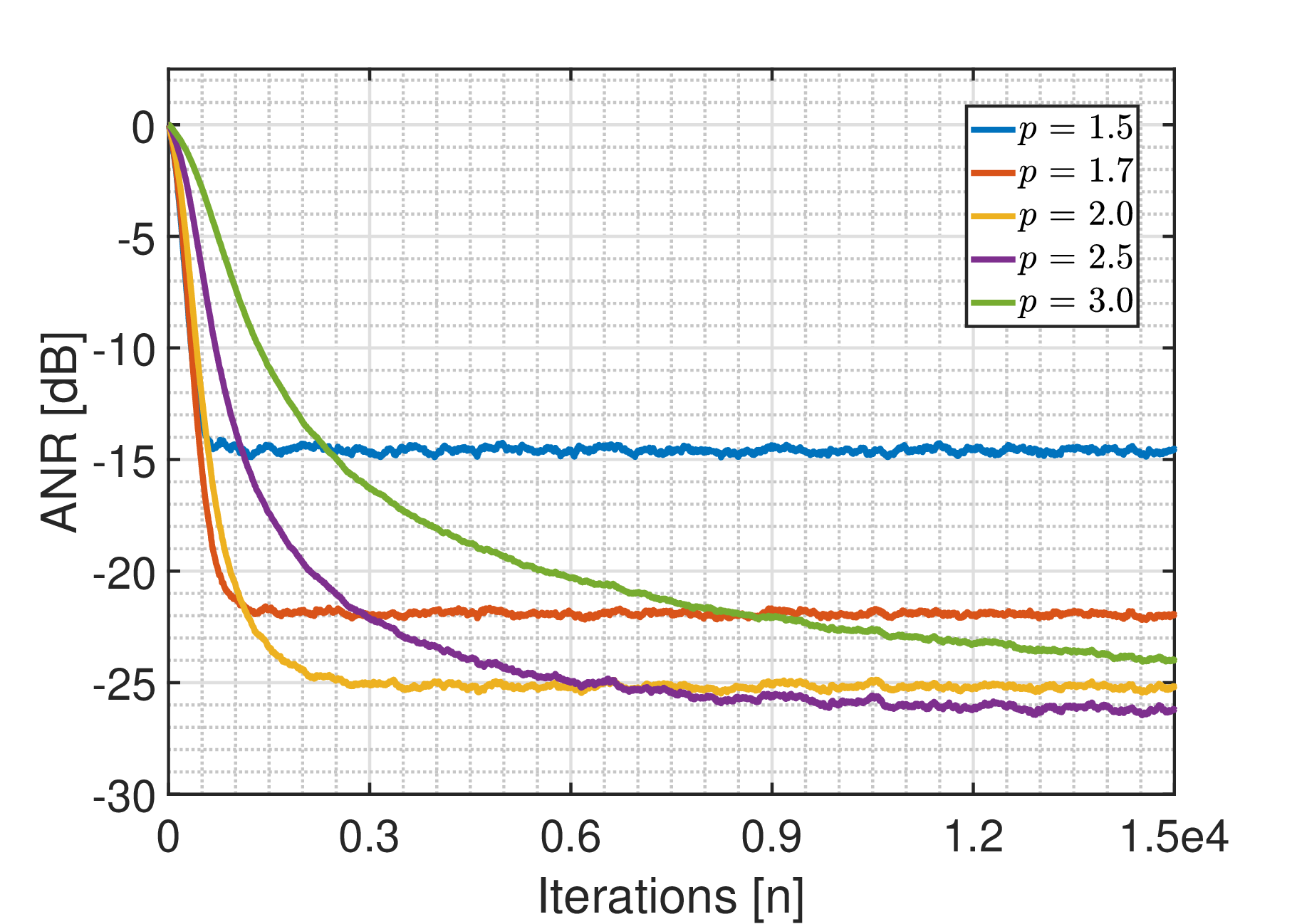}
         \caption{ANR performance at different values of algorithm's parameter $p$ ($\alpha_s$ = 1.5).}
         \label{HEKM_p}
     \end{center}
 \end{figure}

The parameters of the algorithms are {observed accordingly with the studied references cited and} 
chosen after an exhaustive experimental study based on and respecting the convergence and stability analysis demonstrated previously in Section IV. {As follows are presented the performance comparisons of the algorithm varying its main parameters. The Figure \ref{HEKM_alfa} shows the performance of the FXHEKM algorithm changing the value of the parameter $\alpha$. Analogously, Figure \ref{HEKM_p} brings the same evaluation to the different values of the parameter $p$, which represents the exponent value of the HEKM function (Eq.\eqref{dJdw}).}

The values used in this work for each algorithm are FXLMS: $\mu$ = 0.1; FXGMCC: $\mu$ = 0.0495, $\sigma$ = 1.5, $p$ = 1.7 and $\nu$ = 1.0; FXGHT: \textcolor{blue}{$\rho$} = 0.1, $\lambda$ = 0.4, $p$ = 2 and $\sigma$ = 14.5; IFXGMCC: $\mu$ = 0.0535, $\sigma$ = 2.0, p = 1.5 and $\nu$ = 0.5; FXGR: $\mu$ = 0.1 and $\zeta$ = 0.2; FXECH: $\gamma$ = $\exp$(1), $\mu$ = 0.034 and $\lambda$ = 3.4 and $p$ = 2; FXHEKM: \textcolor{blue}{$\rho$} = 0.1, $\eta$ = 1.0, $\alpha$ = 0.4, $\zeta$ = 0.2 and $p$ = 2.

\begin{figure}[ht!]
    \begin{center}
        \includegraphics[scale=.3]{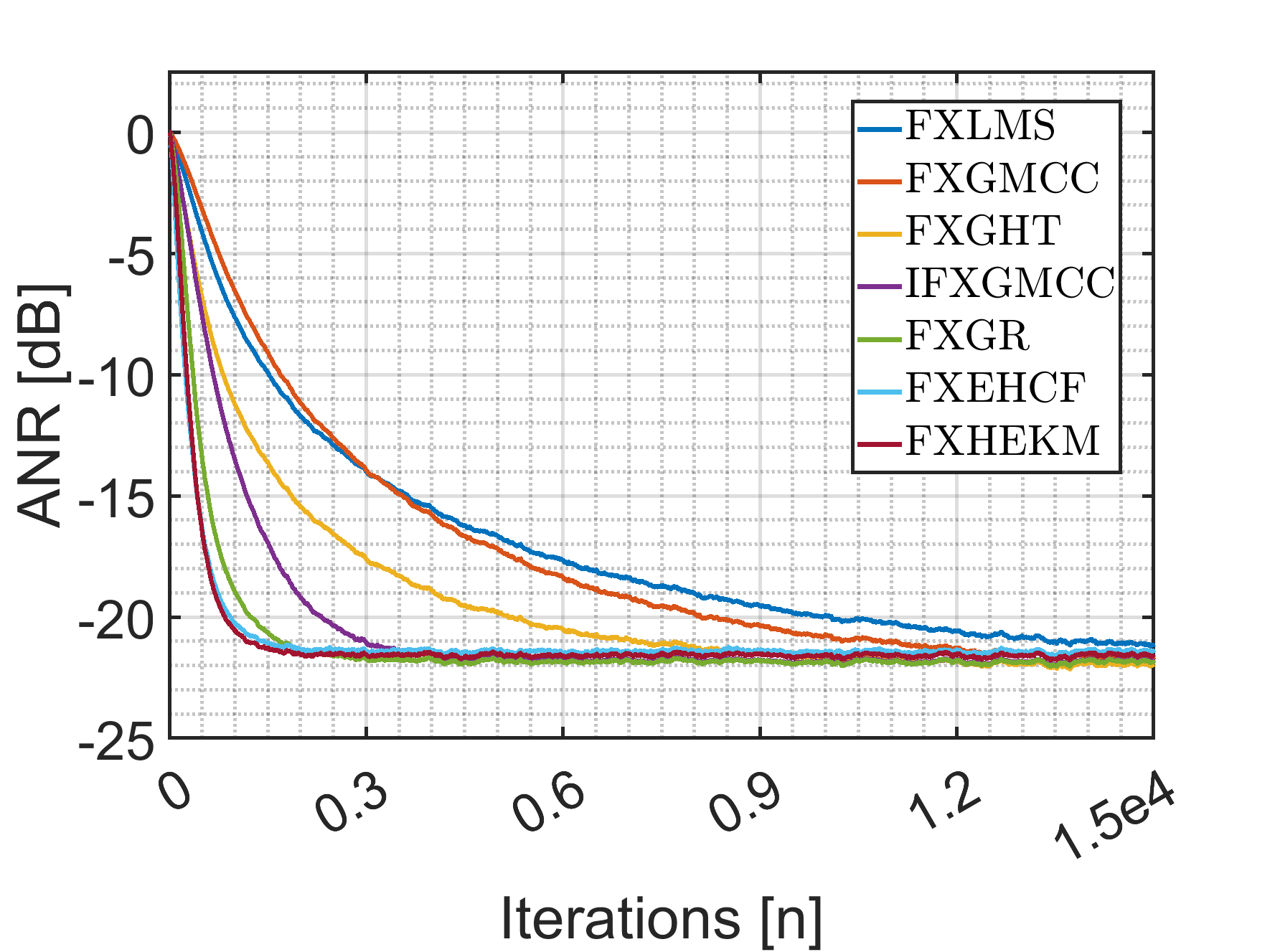}
        \caption{ANR comparison between HEKM method against other classical/robust approaches for Gaussian distribution noise ($\alpha_s$ = 2.0) e SNR = 20 dB.} 
        \label{ANR1_a20}
    \end{center}
\end{figure}

The next example is conducted using the $\alpha$-stable noise modeled as a Gaussian distribution ($\alpha_s=2$). Fig. \ref{ANR1_a20} shows the performance comparison between the different adaptive filters applied in the noise canceling of the white noise input. The curves show that the algorithms can reach a steady state with a similar ANR level for each curve. However, the curve of the proposed FXHEKM algorithm reached the lowest ANR with reasonable convergence speed and close to -22dB in mean.


Another important aspect to note is that the proposed FXHEKM algorithm has a fast response in terms of ANR that reduces the number of required iterations. The proposed FXHEKM algorithm converges much quicker than standard algorithms and other robust approaches with a little improvement. FXHEKM is more robust against impulsive noise and employs an adaptation strategy that learns faster than competing robust algorithms. Moreover, the FXHEKM achieves one of the lowest MSE values during all iterations. 

The combination of the hyperbolic tangent objective curve (similar to GHT \cite{ref9}) with an exponential argument, the last being responsible for increased robustness, combined with a kernel approach (used in FXGMCC and IFXGMCC of \cite{ref8}) provides a simple filtered-X structure robust performance for ANR systems.



\begin{figure}[htbp]
    \centerline{\includegraphics[scale=.23]{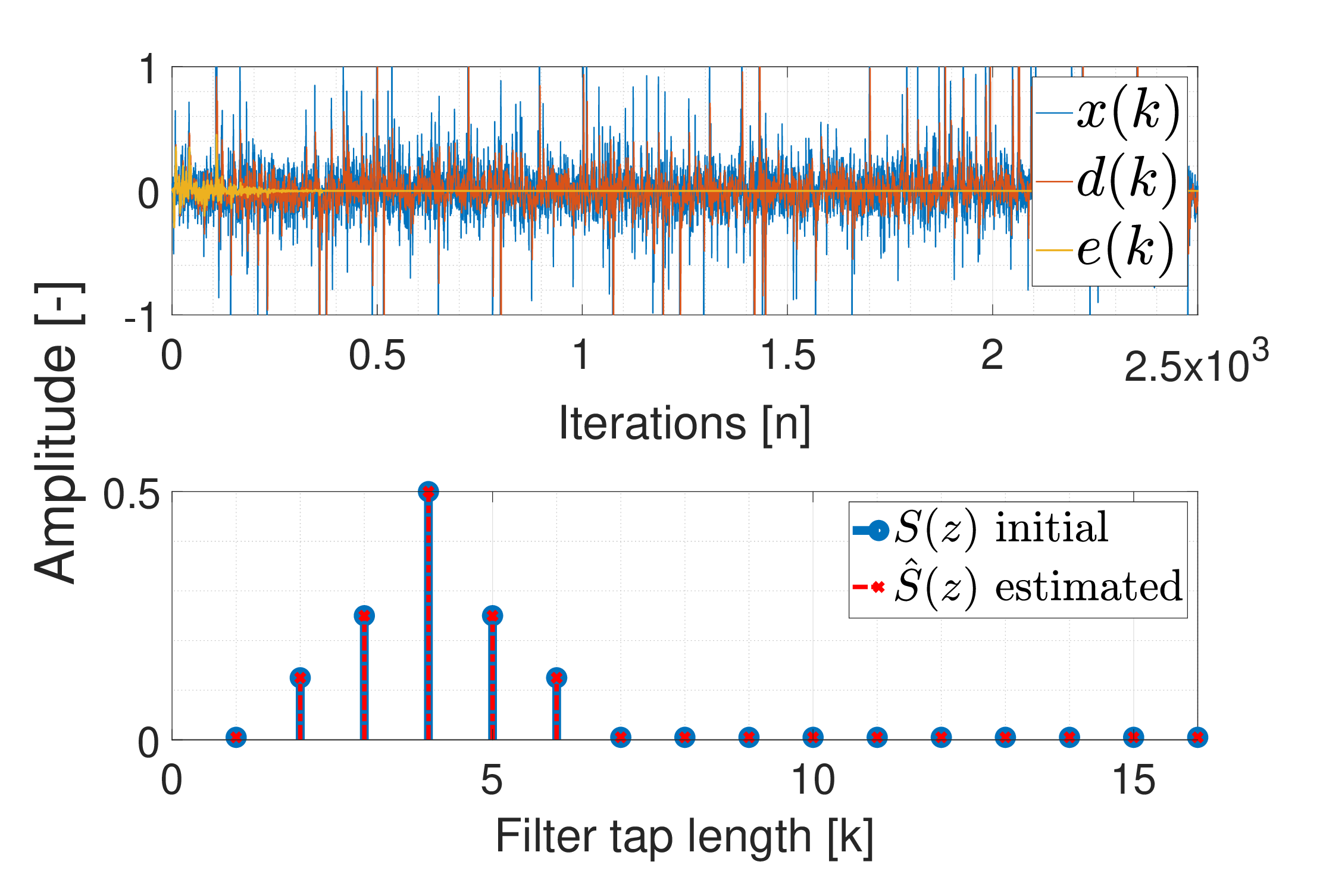}}
    \caption{Analysis of the initial system identification with the estimation of the secondary path $\hat{S}(z)$ to scenario 2.}
    \label{fig2_2}
\end{figure}

In the following experiments, we consider scenario 2, where impulsive noise signals are employed. For these numerical examples, we employ the $\alpha-$stable distribution noise with $\alpha_s<2$ (pseudo-impulsive noise), which contains a signal with impulsive peaks mixed with some Gaussian noise. The idea to evaluate the impulsive noise is based on a range of practical problems \cite{vase}. In the following example, Fig. \ref{fig2_2} shows the performance of the $\hat{S}(z)$ as an input to the ANC system.

\begin{figure}[ht!]
    \begin{center}
    	\includegraphics[scale=.29]{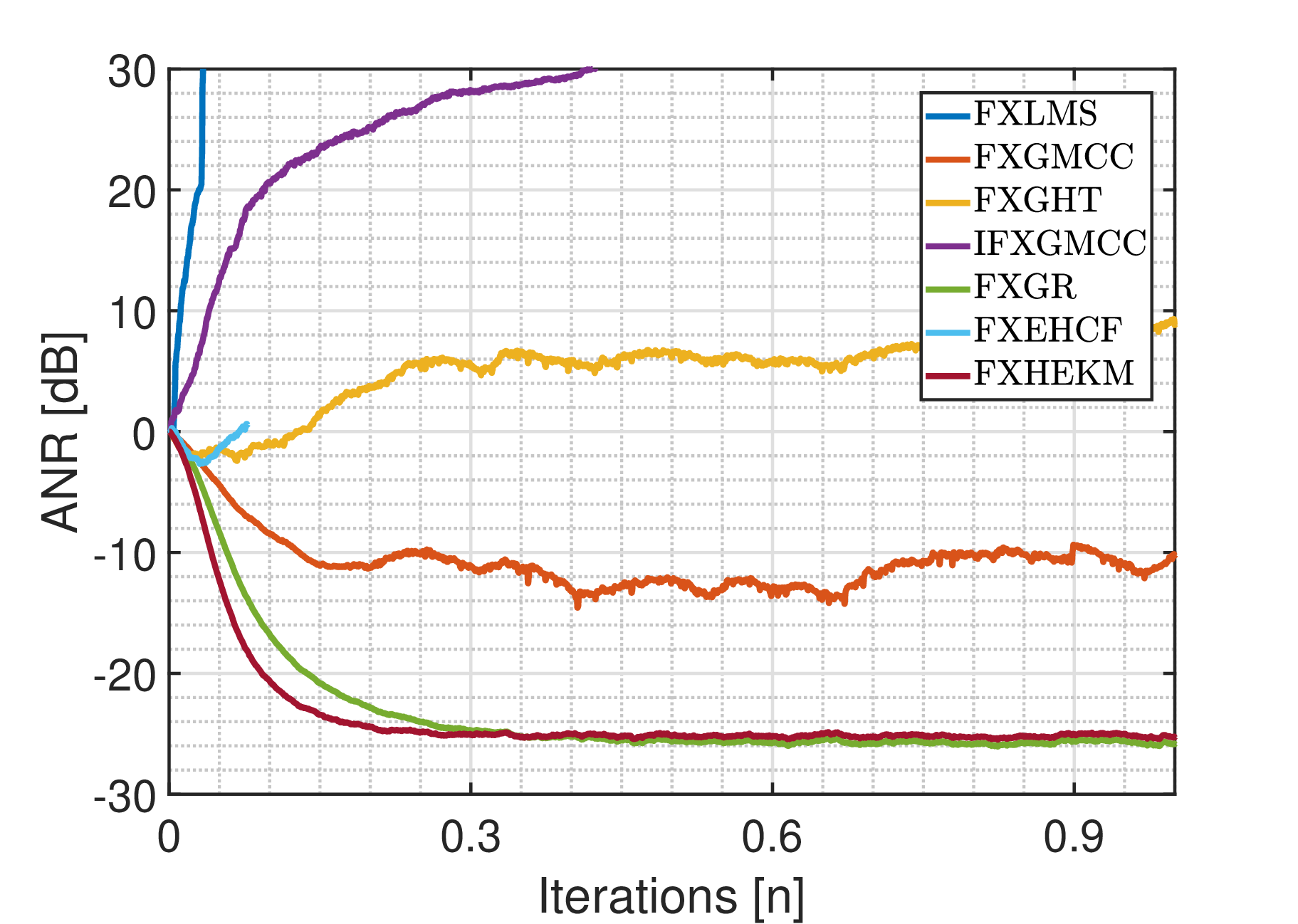}
        \caption{ANR comparison between HEKM method against other classical/robust approaches for impulsive noise ($\alpha_s$ = 1.5) and SNR $\approx$ 25dB.}
        \label{fig2_3}
    \end{center}
\end{figure}

The results above ensure that the ANC system behaves properly in the residual error computation even with this $\alpha$-stable distribution. The following analysis of ANR performance evaluates the proposed ANC framework comparing all studied algorithms in the presence of non-Gaussian noise. Fig. \ref{fig2_3} shows the results of the ANR performance for all the methods compared in this study.


In this example, the ANR performance per iteration is compared again to evaluate the analyzed algorithms.
As expected, the FXLMS algorithm turns out to be unstable (dark blue line, after 500 iterations), and in general, the performance of all algorithms is downgraded, revealing the difficulty of the learning phase due to the non-linear and abrupt characteristics of the impulsive signal.

Nonetheless, it is notable how the FXHEKM behaves with the impulsive noise, which is perceptible for the influence of the M-estimate performance (see the closed performance of FXHEKM and FXGR versions in the second scenario). In contrast, the other objective functions aligned only to the filtered-X structure do not work well in this example, and in some cases (IFXGMCC, FXGHT and FXECHF), the algorithms show an unstable behaviour. 
The proposed FXHEKM maintained the best performance, converging faster  and outperforming the FXGR algorithm under impulsive noise scenarios for more than $3$dB until $n = 2000$ iterations. After that, both approaches have very close performance at the steady state.





The simulations performed to set the parameters of all methods confirm the hypothesis presented about the performance of the proposed FXHEKM algorithm. As shown in the plot, the FXHEKM algorithm reaches the best performance, outperforming the other methods.



\begin{figure}[ht!]
   \begin{center}
\includegraphics[width=1.\columnwidth]{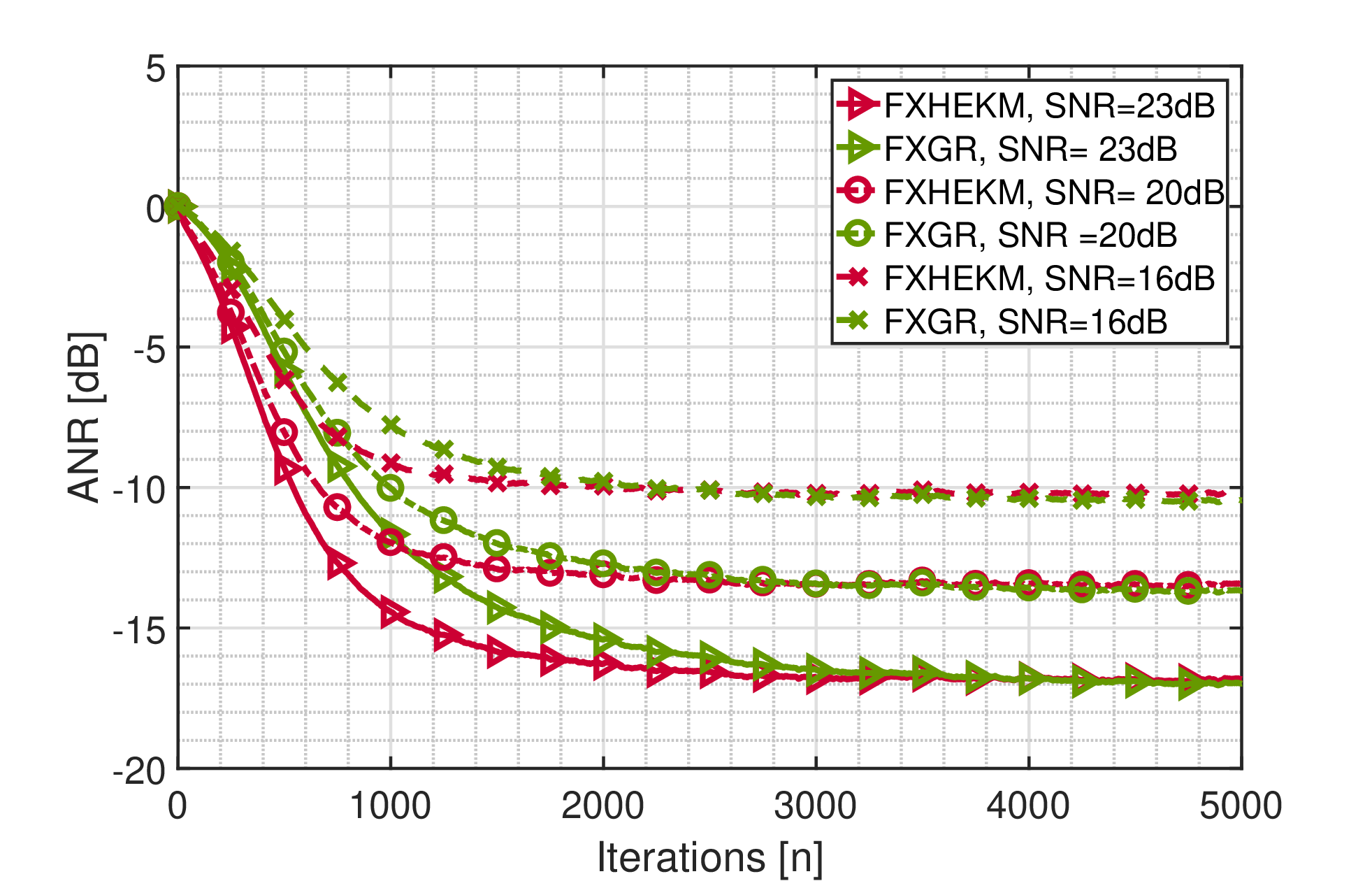}
       \caption{ANR performance at different SNR levels.} 
       \label{ANR1_HEKM}
   \end{center}
\end{figure}

Next, we compare the performance of the two algorithms that worked well against impulsive noise, namely, the FXHEKM and the FXGR algorithms. For this analysis, we consider a total of $1000$ Monte Carlo simulations. Fig. \ref{ANR1_HEKM} depicts the performance of both algorithms for different SNR levels. The proposed FXHEKM shows a much faster convergence speed than the FXGR algorithm for all SNR considered, while achieving a similar performance in terms of ANR at steady state.



\begin{figure}[ht!]
     \begin{center}
     	\includegraphics[width=1.\columnwidth]{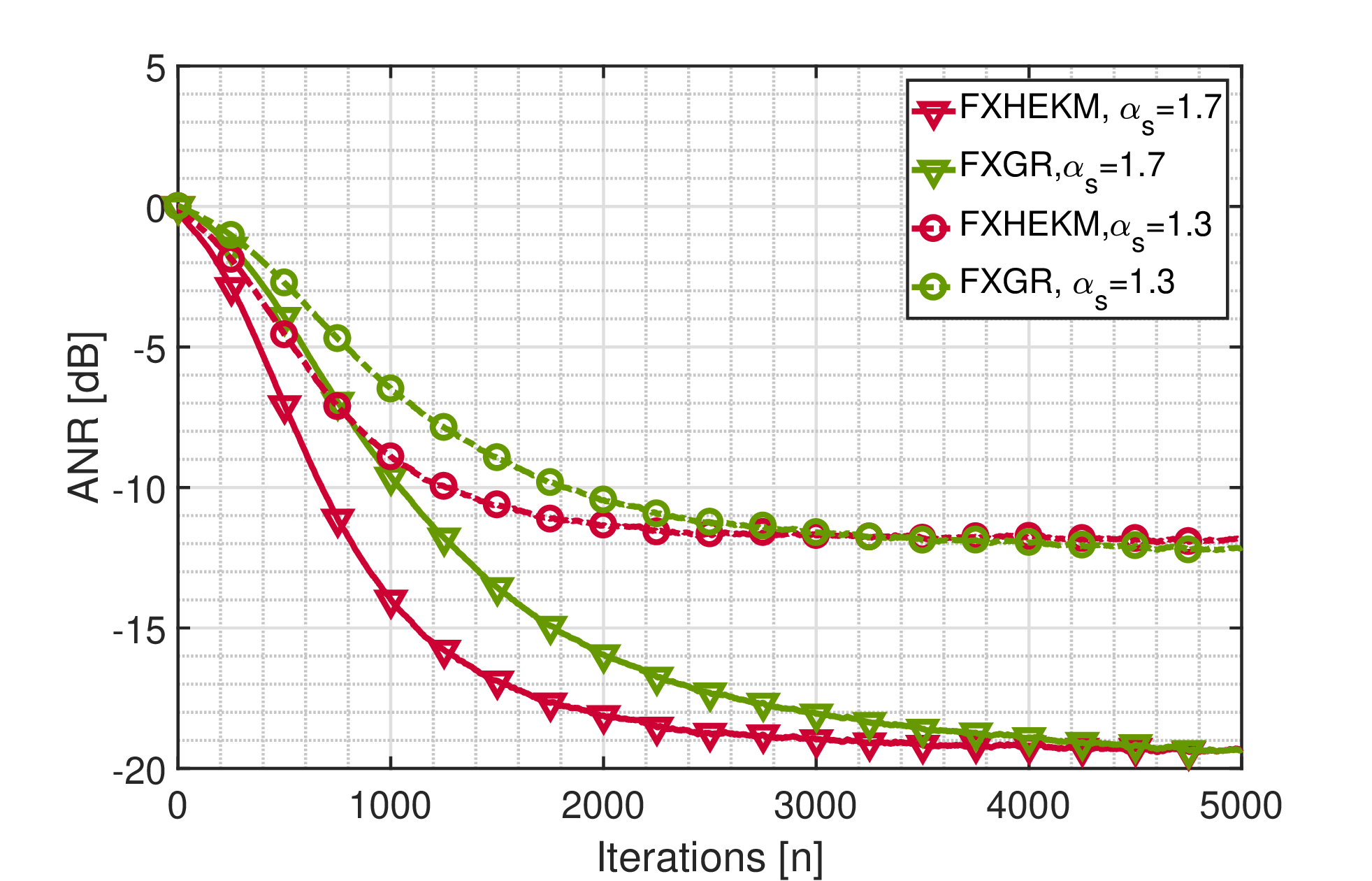}
         \caption{ANR performance at different impulsive $\alpha_s$ values (SNR $\approx$ 23dB).}
         \label{ANR2_HEKM}
     \end{center}
 \end{figure}

Finally, we tested the execution of the FXGR and FXHEKM algorithms with different values of $\alpha_s$. Specifically, we considered a softer impulsive noise ($\alpha_s = 1.7$) and an extremely hard impulsive noise ($\alpha_s = 1.3$). The results of this experiment are shown in Fig. \ref{ANR2_HEKM}. 

 Lower values of $\alpha_s$ degrade the ANR performance of the algorithms. However, both algorithms converge, even under the extreme case of $\alpha_s=1.3$. Moreover, the proposed FXHEKM algorithm achieves the fastest convergence for all values of $\alpha_s$, while obtaining a similar ANR performance when compared to the FXGR algorithm, which demonstrates the efficiency and robustness of the proposed technique.  The FXHEKM algorithm shows stability and fast convergence, as well as adaptability to different ANC simulations. Indeed, the proposed FXHEKM algorithm outperforms the other algorithms in the scenarios studied, suggesting the interesting feasibility for other applications and future studies.


\section{Conclusions}

In this work, we have proposed a new robust adaptive algorithm for ANC, which was denoted FXHEKM. In particular, the proposed FXHEKM can deal with noise characterized by an $\alpha-$stable distribution. A statistical analysis of the proposed FXHEKM algorithm is carried out along with a study of its computational cost. Numerical results have shown the cost-effectiveness of the proposed FXHEKM algorithm to cancel the presence of additive spurious signals, such as $\alpha$-stable noises, against competing algorithms. {Since this work focused on robust ANC algorithms we opted for using Gaussian and non-Gaussian noise examples. Future work might consider other types of noise such as traffic noise and factory noise.}

\bibliographystyle{IEEEtran}
\bibliography{jbib}

\begin{thebibliography}{10}
\providecommand{\url}[1]{#1}
\csname url@samestyle\endcsname
\providecommand{\newblock}{\relax}
\providecommand{\bibinfo}[2]{#2}
\providecommand{\BIBentrySTDinterwordspacing}{\spaceskip=0pt\relax}
\providecommand{\BIBentryALTinterwordstretchfactor}{4}
\providecommand{\BIBentryALTinterwordspacing}{\spaceskip=\fontdimen2\font plus
\BIBentryALTinterwordstretchfactor\fontdimen3\font minus \fontdimen4\font\relax}
\providecommand{\BIBforeignlanguage}[2]{{%
\expandafter\ifx\csname l@#1\endcsname\relax
\typeout{** WARNING: IEEEtran.bst: No hyphenation pattern has been}%
\typeout{** loaded for the language `#1'. Using the pattern for}%
\typeout{** the default language instead.}%
\else
\language=\csname l@#1\endcsname
\fi
#2}}
\providecommand{\BIBdecl}{\relax}
\BIBdecl

\bibitem{ref1}
H.~Olson, ``Electronic control of noise, vibration, and reverberation,'' \emph{The Journal of The Acoustical Society of America}, vol.~28, no.~5, 1956.

\bibitem{ref25}
B.~Widrow and S.~D. Stearns, \emph{Adaptive Signal Processing}.\hskip 1em plus 0.5em minus 0.4em\relax Prentice-Hall, 1985.

\bibitem{ref4}
S.~Kuo and D.~Morgan, ``Active noise control: a tutorial review,'' \emph{Proceedings of the IEEE}, vol.~87, no.~6, pp. 943--973, 1999.

\bibitem{ref2}
B.~Widrow, J.~Glover, J.~McCool, J.~Kaunitz, C.~Williams, R.~Hearn, J.~Zeidler, J.~Eugene~Dong, and R.~Goodlin, ``Adaptive noise cancelling: Principles and applications,'' \emph{Proceedings of the IEEE}, vol.~63, no.~12, pp. 1692--1716, 1975.

\bibitem{aifir}
R.~de~Lamare and R.~Sampaio-Neto, ``Adaptive reduced-rank mmse filtering with interpolated fir filters and adaptive interpolators,'' \emph{IEEE Signal Processing Letters}, vol.~12, no.~3, pp. 177--180, 2005.

\bibitem{jio}
R.~C. de~Lamare and R.~Sampaio-Neto, ``Reduced-rank adaptive filtering based on joint iterative optimization of adaptive filters,'' \emph{IEEE Signal Processing Letters}, vol.~14, no.~12, pp. 980--983, 2007.

\bibitem{jiodoa}
L.~Wang, R.~C. de~Lamare, and M.~Haardt, ``Direction finding algorithms based on joint iterative subspace optimization,'' \emph{IEEE Transactions on Aerospace and Electronic Systems}, vol.~50, no.~4, pp. 2541--2553, 2014.

\bibitem{jiomimo}
R.~C. de~Lamare and R.~Sampaio-Neto, ``Adaptive reduced-rank equalization algorithms based on alternating optimization design techniques for mimo systems,'' \emph{IEEE Transactions on Vehicular Technology}, vol.~60, no.~6, pp. 2482--2494, 2011.

\bibitem{jiols}
------, ``Reduced-rank space–time adaptive interference suppression with joint iterative least squares algorithms for spread-spectrum systems,'' \emph{IEEE Transactions on Vehicular Technology}, vol.~59, no.~3, pp. 1217--1228, 2010.

\bibitem{saalt}
------, ``Sparsity-aware adaptive algorithms based on alternating optimization and shrinkage,'' \emph{IEEE Signal Processing Letters}, vol.~21, no.~2, pp. 225--229, 2014.

\bibitem{dce}
S.~Xu, R.~C. de~Lamare, and H.~V. Poor, ``Distributed compressed estimation based on compressive sensing,'' \emph{IEEE Signal Processing Letters}, vol.~22, no.~9, pp. 1311--1315, 2015.

\bibitem{jidf}
R.~C. de~Lamare and R.~Sampaio-Neto, ``Adaptive reduced-rank processing based on joint and iterative interpolation, decimation, and filtering,'' \emph{IEEE Transactions on Signal Processing}, vol.~57, no.~7, pp. 2503--2514, 2009.

\bibitem{rrmser}
Y.~Cai, R.~C. de~Lamare, B.~Champagne, B.~Qin, and M.~Zhao, ``Adaptive reduced-rank receive processing based on minimum symbol-error-rate criterion for large-scale multiple-antenna systems,'' \emph{IEEE Transactions on Communications}, vol.~63, no.~11, pp. 4185--4201, 2015.

\bibitem{rrlrd}
L.~Qiu, Y.~Cai, R.~C. de~Lamare, and M.~Zhao, ``Reduced-rank doa estimation algorithms based on alternating low-rank decomposition,'' \emph{IEEE Signal Processing Letters}, vol.~23, no.~5, pp. 565--569, 2016.

\bibitem{dynovs}
Z.~Shao, L.~T.~N. Landau, and R.~C. de~Lamare, ``Dynamic oversampling for 1-bit adcs in large-scale multiple-antenna systems,'' \emph{IEEE Transactions on Communications}, vol.~69, no.~5, pp. 3423--3435, 2021.

\bibitem{ref5}
P.~Nelson and S.~Elliot, \emph{Active Control of Sound}.\hskip 1em plus 0.5em minus 0.4em\relax Academic Press, 1992.

\bibitem{ref8}
Y.-R. Chien, C.-H. Yu, and H.-W. Tsao, ``Affine-projection-like maximum correntropy criteria algorithm for robust active noise control,'' \emph{IEEE/ACM Transactions on Audio, Speech, and Language Processing}, vol.~30, pp. 2255--2266, 2022.

\bibitem{ref17}
E.~Bjarnason, ``Analysis of the filtered-{X} {LMS} algorithm,'' \emph{IEEE Transactions on Speech and Audio Processing}, vol.~3, no.~6, pp. 504--514, 1995.

\bibitem{ref6}
L.~Lu, K.-L. Yin, R.~C. {de Lamare}, Z.~Zheng, Y.~Yu, X.~Yang, and B.~Chen, ``A survey on active noise control in the past decade—part {I}: Linear systems,'' \emph{Signal Processing}, vol. 183, 2021.

\bibitem{ref7}
------, ``A survey on active noise control in the past decade–part {II}: Nonlinear systems,'' \emph{Signal Processing}, vol. 181, 2021.

\bibitem{ref21}
B.~Huang, Y.~Xiao, J.~Sun, and G.~Wei, ``A variable step-size {FXLMS} algorithm for narrowband active noise control,'' \emph{IEEE Transactions on Audio, Speech, and Language Processing}, vol.~21, no.~2, pp. 301--312, 2013.

\bibitem{ref22}
A.~Zeb, A.~Mirza, Q.~U. Khan, and S.~A. Sheikh, ``Improving performance of {FxRLS} algorithm for active noise control of impulsive noise,'' \emph{Applied Acoustics}, vol. 116, pp. 364--374, 2017.

\bibitem{ref18}
K.-L. Yin, Y.-F. Pu, and L.~Lu, ``Robust {Q}-gradient subband adaptive filter for nonlinear active noise control,'' \emph{IEEE/ACM Transactions on Audio, Speech, and Language Processing}, vol.~29, pp. 2741--2752, 2021.

\bibitem{ref23}
M.~T. Akhtar and W.~Mitsuhashi, ``Improving robustness of filtered-x least mean p-power algorithm for active attenuation of standard symmetric-$\alpha$-stable impulsive noise,'' \emph{Applied Acoustics}, vol.~72, no.~9, pp. 688--694, 2011.

\bibitem{ref24}
N.~K. Rout, D.~P. Das, and G.~Panda, ``Particle swarm optimization based nonlinear active noise control under saturation nonlinearity,'' \emph{Applied Soft Computing}, vol.~41, pp. 275--289, 2016.

\bibitem{l1stap}
Z.~Yang, R.~C. de~Lamare, and X.~Li, ``$l_1$ -regularized stap algorithms with a generalized sidelobe canceler architecture for airborne radar,'' \emph{IEEE Transactions on Signal Processing}, vol.~60, no.~2, pp. 674--686, 2012.

\bibitem{locsme}
H.~Ruan and R.~C. de~Lamare, ``Robust adaptive beamforming using a low-complexity shrinkage-based mismatch estimation algorithm,'' \emph{IEEE Signal Processing Letters}, vol.~21, no.~1, pp. 60--64, 2014.

\bibitem{okspme}
------, ``Robust adaptive beamforming based on low-rank and cross-correlation techniques,'' \emph{IEEE Transactions on Signal Processing}, vol.~64, no.~15, pp. 3919--3932, 2016.

\bibitem{lrcc}
------, ``Distributed robust beamforming based on low-rank and cross-correlation techniques: Design and analysis,'' \emph{IEEE Transactions on Signal Processing}, vol.~67, no.~24, pp. 6411--6423, 2019.

\bibitem{rapa}
A.~R. Flores and R.~C. de~Lamare, ``Robust and adaptive power allocation techniques for rate splitting based mu-mimo systems,'' \emph{IEEE Transactions on Communications}, vol.~70, no.~7, pp. 4656--4670, 2022.

\bibitem{WANG2025}
\BIBentryALTinterwordspacing
Y.~Wang, B.~Lin, Y.~Guan, J.~Qian, Y.-R. Chien, and G.~Qian, ``Fractional-order generalized complex correntropy algorithm for robust active noise control,'' \emph{Signal Processing}, vol. 235, p. 110024, 2025. [Online]. Available: \url{https://www.sciencedirect.com/science/article/pii/S0165168425001380}
\BIBentrySTDinterwordspacing

\bibitem{ref9}
Y.~Xiao, S.~Chen, Q.~Zhang, D.~Lin, M.~Shen, J.~Qian, and S.~Wang, ``Generalized hyperbolic tangent based random fourier conjugate gradient filter for nonlinear active noise control,'' \emph{IEEE/ACM Transactions on Audio, Speech, and Language Processing}, vol.~31, pp. 619--632, 2023.

\bibitem{Ye2023}
J.~Ye, Y.~Yu, B.~Chen, and Z.~Zheng, ``Optimal subband adaptive filtering algorithm over functional link neural network,'' in \emph{IEEE 33rd International Workshop on Machine Learning for Signal Processing (MLSP)}, 2023, pp. 1--6.

\bibitem{Ye2024}
J.~Ye, Y.~Yu, B.~Chen, Z.~Zheng, and J.~Chen, ``Optimal subband adaptive filter over functional link neural network: Algorithms and applications,'' \emph{IEEE Transactions on Circuits and Systems I: Regular Papers}, pp. 1--14, 2024.

\bibitem{ref10}
K.~Kumar, R.~Pandey, S.~S. Bhattacharjee, and N.~V. George, ``Exponential hyperbolic cosine robust adaptive filters for audio signal processing,'' \emph{IEEE Signal Processing Letters}, vol.~28, pp. 1410--1414, 2021.

\bibitem{deepanc1}
Z.~Luo, D.~Shi, and W.-S. Gan, ``A hybrid {SFANC-FxNLMS} algorithm for active noise control based on deep learning,'' \emph{IEEE Signal Processing Letters}, vol.~29, pp. 1102--1106, 2022.

\bibitem{deepanc2}
D.~Chen, L.~Cheng, D.~Yao, J.~Li, and Y.~Yan, ``A secondary path-decoupled active noise control algorithm based on deep learning,'' \emph{IEEE Signal Processing Letters}, vol.~29, pp. 234--238, 2022.

\bibitem{deepanc3}
Z.~Luo, D.~Shi, W.-S. Gan, and Q.~Huang, ``Delayless generative fixed-filter active noise control based on deep learning and bayesian filter,'' \emph{IEEE/ACM Transactions on Audio, Speech, and Language Processing}, vol.~32, pp. 1048--1060, 2024.

\bibitem{ref11}
Y.~Yu, H.~He, R.~C. de~Lamare, and B.~Chen, ``General robust subband adaptive filtering: Algorithms and applications,'' \emph{IEEE/ACM Transactions on Audio, Speech, and Language Processing}, vol.~30, pp. 2128--2140, 2022.

\bibitem{dme}
Y.~Yu, H.~He, T.~Yang, X.~Wang, and R.~C. de~Lamare, ``Diffusion normalized least mean {M}-estimate algorithms: Design and performance analysis,'' \emph{IEEE Transactions on Signal Processing}, vol.~68, pp. 2199--2214, 2020.

\bibitem{Kim2024}
I.~K. de~S.~Hermont, A.~R. Flores, and R.~C. de~Lamare, ``Robust adaptive filtering based on the hyperbolic tangent exponential kernel {M}-estimator function for active noise cancellation,'' in \emph{2024 19th International Symposium on Wireless Communication Systems (ISWCS)}, 2024, pp. 1--6.

\bibitem{ref13}
A.~H. Sayed, \emph{Adaptive Filters}.\hskip 1em plus 0.5em minus 0.4em\relax Wiley, 2008.

\bibitem{yang}
J.~Yang, F.;~Guo and J.~Yang, ``Stochastic analysis of the filtered-x lms algorithm for active noise control,'' \emph{IEEE/ACM Transactions on Audio, Speech, and Language Processing}, vol.~28, pp. 2252--2266, 2020.

\bibitem{ref16}
S.~Kuo and D.~Vijayan, ``A secondary path modeling technique for active noise control systems,'' \emph{IEEE Transactions on Speech and Audio Processing}, vol.~5, no.~4, pp. 374--377, 1997.

\bibitem{vase}
S.~V. Vaseghi, \emph{Advanced Digital Signal Processing and Noise Reduction}.\hskip 1em plus 0.5em minus 0.4em\relax Wiley, 2008.

\end{thebibliography}

\end{document}